\documentclass[sigconf]{acmart}

\usepackage{wrapfig,lipsum,booktabs}
\usepackage{pifont}
\usepackage{microtype}
\usepackage{graphicx}
\usepackage{subfigure}
\usepackage{booktabs} 
\usepackage{paralist}
\usepackage{amsmath}
\usepackage{empheq}
\usepackage{mdframed}
\usepackage[utf8]{inputenc}
\usepackage[english]{babel}
\usepackage{amsthm} 
\usepackage{xspace}
\usepackage{bm}
\usepackage{multirow}

\usepackage{enumitem}
\usepackage{xcolor}
\setlist[itemize]{leftmargin=*}
\usepackage{algorithm}
\usepackage{algpseudocode}
\usepackage{graphicx}

\usepackage{color, colortbl}
\usepackage{cellspace}
\newcommand\scalemath[2]{\scalebox{#1}{\mbox{\ensuremath{\displaystyle #2}}}}

\theoremstyle{definition}

\newcommand{\hide}[1]{}

\makeatletter
\newcommand\footnoteref[1]{\protected@xdef\@thefnmark{\ref{#1}}\@footnotemark}
\makeatother

\renewcommand{\algorithmicrequire}{\textbf{Input:}}
\renewcommand{\algorithmicensure}{\textbf{Output:}}
\renewcommand{\algorithmiccomment}[1]{\hfill$\blacktriangleright$ #1}

\newcommand{\cbit}{\begin{compactitem}}
\newcommand{\ceit}{\end{compactitem}}
\newcommand{\cben}{\begin{compactenum}}
\newcommand{\ceen}{\end{compactenum}}

\newcommand{\beq}{\begin{equation}}
	\newcommand{\eeq}{\end{equation}}

\newcommand{\beqn}{\begin{equation*}}
\newcommand{\eeqn}{\end{equation*}}

\newcommand{\bit}{\begin{itemize}}
	\newcommand{\eit}{\end{itemize}}
\newcommand{\ben}{\begin{enumerate}}
	\newcommand{\een}{\end{enumerate}}

\newcounter{x}

\newcommand{\mF}{\mathcal{F}}

\newcommand{\mH}{\mathcal{H}}

\newcommand{\mC}{\mathcal{C}}

\newcommand{\bdel}{\boldsymbol{\Delta}}

\newcommand{\bfl}{\mathbf{f}}
\newcommand{\bx}{\mathbf{x}}

\newcommand{\br}{\mathbf{r}}
\newcommand{\bs}{\mathbf{s}}
\newcommand{\bzb}{\bar{\mathbf{z}}}
\newcommand{\bz}{\mathbf{z}}
\newcommand{\bw}{\mathbf{w}}

\newcommand{\bpi}{\bm{\pi}}

\newcommand{\be}{\bm{\epsilon}}
\newcommand{\bA}{\mathbf{A}}

\newcommand{\R}{\mathbb{R}}

\newcommand{\Z}{\mathbb{Z}}

\newcommand{\method}{{\sc ALARM}\xspace}
\newcommand{\ex}{{\sc xStream}\xspace}

\newcommand{\fin}{{\sc Czech}\xspace}
\newcommand{\card}{{\sc Card}\xspace}

\newcommand{\clus}{{\sc Sum}\xspace}
\newcommand{\expl}{{\sc Expl}\xspace}
\newcommand{\cand}{{\sc Cand}\xspace}
\newcommand{\rdi}{{\sc RDI}\xspace}

\newcommand{\la}[1]{{\color{blue}#1}}

\AtBeginDocument{%
	\providecommand\BibTeX{{%
			\normalfont B\kern-0.5em{\scshape i\kern-0.25em b}\kern-0.8em\TeX}}}

\setcopyright{acmcopyright}
\copyrightyear{2023}
\acmYear{2023}
\acmConference[KDD '23]{Proceedings of the 29th ACM SIGKDD Conference on Knowledge Discovery and Data Mining}{August 6--10, 2023}{Long Beach, CA, USA}
\acmPrice{15.00}

\begin{document}

\title[An End-to-End Human-in-the-loop Framework for Anomaly Reasoning and Management]{From Explanation to Action: An End-to-End Human-in-the-loop Framework for Anomaly Reasoning and Management} %



\author{Xueying Ding}
\affiliation{%
  \institution{Carnegie Mellon University}
  \institution{Heinz College of Information Systems}
  \country{}
}
\email{xding2@andrew.cmu.edu}

\author{Nikita Seleznev}
\affiliation{%
	\institution{Capital One}
	\institution{Center for Machine Learning}
	\country{}
}
\email{nikita.seleznev@capitalone.com}

\author{Senthil Kumar}
\affiliation{%
	\institution{Capital One}
	\institution{Center for Machine Learning}
	\country{}
}
\email{senthil.kumar@capitalone.com}

\author{C. Bayan Bruss}
\affiliation{%
	\institution{Capital One}
	\institution{Center for Machine Learning}
	\country{}
}
\email{bayan.bruss@capitalone.com}

\author{Leman Akoglu}
\affiliation{%
	\institution{Carnegie Mellon University}
	\institution{Heinz College of Information Systems}
	\country{}
}
\email{lakoglu@andrew.cmu.edu}

\begin{abstract}
Anomalies are often indicators of malfunction or inefficiency in various  systems such as manufacturing, healthcare, finance, surveillance, to name a few. While the literature is abundant in effective detection algorithms due to this practical relevance, autonomous anomaly detection is rarely used in real-world scenarios. Especially in high-stakes applications, a human-in-the-loop is often involved in  processes \textit{beyond} detection such as 
verification and troubleshooting.
In this work, we introduce {\sc ALARM}\textsuperscript{\ref{note1}} (for \underline{\textsc{A}}nalyst-in-the-\underline{\textsc{L}}oop \underline{\textsc{A}}nomaly \underline{\textsc{R}}easoning and \underline{\textsc{M}}anagement); an end-to-end framework that supports the anomaly mining cycle comprehensively, from detection to action. 
Besides 
unsupervised detection of emerging anomalies, it offers anomaly explanations and an interactive GUI for human-in-the-loop processes---visual exploration, sense-making, and ultimately action-taking via designing new detection rules---that help close ``the loop'' as the new rules 
complement rule-based supervised detection, typical of many deployed systems in practice.
We demonstrate \method's efficacy
through a series of case studies with fraud analysts from the financial industry.

\end{abstract}

%

\vspace{-0.1in}

\maketitle

\section{Introduction}
\label{sec:intro}

Anomalies appear in many real-world domains,  often as indicators of fault, inefficiency or malfunction in various systems such as manufacturing, environmental monitoring, surveillance, finance, computer networks, to name a few. Therefore, a large body of literature has been devoted to outlier detection algorithms \cite{chandola2009anomaly,han2012outlier,books/sp/Aggarwal2013} as well as open-source tools \cite{achtert2010visual,JMLRpyod,hundman2018detecting,lai2021tods}.

Despite effective outlier detection algorithms, autonomous anomaly detection systems are rarely used in real world scenarios as off-the-shelf algorithms do not work well in complex situations \cite{riveiro2009reasoning}.
The reason is that anomaly detection is an under-specified problem and statistical outliers are not always semantically relevant \cite{sommer2010outside}. Therefore, fully automatic approaches are often impractical and the human expert (or analyst) participation and intervention are crucial.
Especially in high-stakes applications, it is required, often as part of mandated policies, that  the detected anomalies (e.g. hospitals flagged as fraudulent, or credit card users flagged as malicious) go through an auditing process where the human-in-the-loop reasons, validates and troubleshoots these anomalous instances.

\begin{figure*}[!ht]
	\begin{center}
\includegraphics[width=0.99\linewidth]{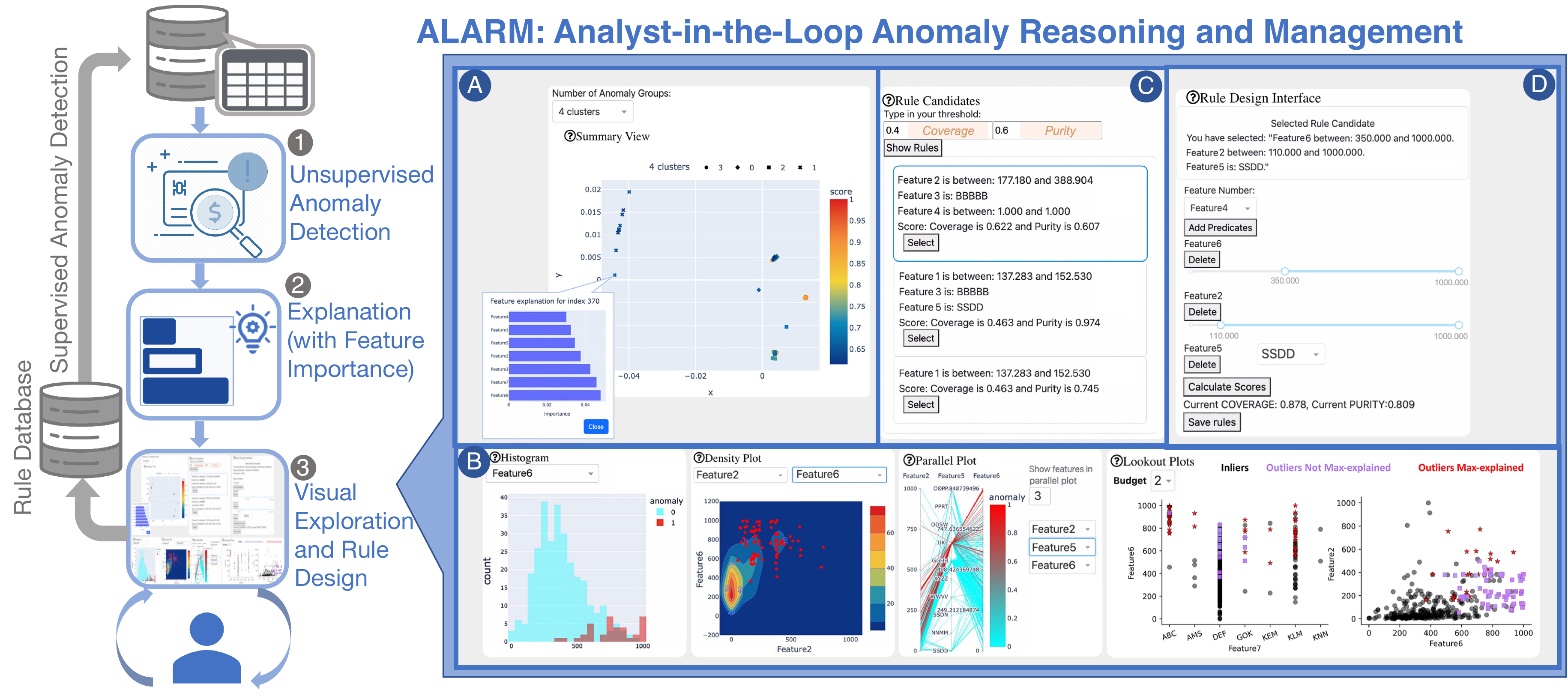}
\end{center}
\vspace{-0.15in}
 \caption{ (left) The series of steps that the proposed \method framework supports for end-to-end anomaly discovery and management. (right) A screenshot of \method's GUI for visual exploration, verification and action-taking (i.e. new rule design). \label{fig:sshot}}
		\vspace{-0.15in}
\end{figure*}
{\bf Motivation.~}
Relative to the vast body of existing work on anomaly detection, the literature is significantly scarce on \textit{post}-detection, \textit{human-in-the-loop} steps of the anomaly mining process.
The lack of support for a full pipeline involving all the steps of the process renders the applicability to real world scenarios inadequate. 

In the last few years, anomaly explanation has gained some attention, with the aim to equip the human analyst with the understanding of why the detected anomalies stand out \cite{panjei2022survey}.
Stand-alone explanations, however, are typically not directly utilized to improve downstream steps. 
There also exist various visual analytics tools specifically developed to aid detection by human perception or visual inspection to aid verification \cite{shi2020visual,ko2016survey,shiravi2011survey}. (See Sec. \ref{sec:related} for related work.)
However, while the explanations and visualizations are expected to help the analyst gain sufficient ``insight'' into the nature of the anomalies, with the hope that they will be able to take suitable action, they lack explicit guidance toward decision-making and action-taking.
Moreover, these detection, explanation and visualization techniques are often developed separately rather than supporting an end-to-end pipeline for human-in-the-loop anomaly mining and management for real world applications.

{\bf Application Scenario.~}
Motivated by these gaps in the literature, we propose an end-to-end framework for anomaly mining, reasoning and management that not only equips the human-in-the-loop with anomaly explanations but also puts these explanations to use toward guiding the analyst in action-taking.
Our work is driven by its applications in finance (related to bank/credit/merchant fraud and money-laundering detection and management), yet it can easily be utilized in other domains in which anomaly mining is critical.

Specifically, as shown in Fig. \ref{fig:sshot}(left), we envision a deployed system where the incoming (in our case, transaction) data stream is screened through a database (i.e. ensemble) of anomaly detection rules for flagging the \textit{known} type of anomalies. Rule-based detection is quite common in many real world deployed systems, thanks to the simplicity and transparency of rules (a small set of feature predicates), fast inference time, and ability to design and deploy new rules in a decentralized fashion by several experts and analysts.
The overarching goal here is to quickly detect and verify new, \textit{emerging} fraudulent activities and design and deploy new rule(s) that can automatically detect similar fraud  in the near and far future.

{\bf Our Work.~}
Toward this goal, we put forth the following pipeline of components. 
($1$) {\bf Unsupervised Detection:} Besides rule-based supervised detection,  the data is also passed  through an  unsupervised detection algorithm, namely our \ex \cite{manzoor2018xstream,sparx22}, for spotting emerging, \textit{unknown} anomalies. ($2$) {\bf Explanation:} 
We develop a built-in, model-specific explanation algorithm for \ex that
estimates feature importance weights, reflective of subspaces in which the anomalies stand out the most.
Importantly, the value of explanations depend on how humans put them into use \cite{kaur2020interpreting} and to the
extent that they are useful for humans in improving a downstream task
\cite{shen2020useful,jesus2021can}. 
($3$) {\bf Visual Exploration and Rule Design, with Human Interaction:} To this end, as Fig. \ref{fig:sshot}(right) illustrates, we leverage the explanations to present discovered anomalous patterns (i.e. clusters) to the analyst through an interactive visual interface (inset \textbf{A}).
Anomaly clusters (rather than one-off outliers) indicate repeating cases, of which the analyst is interested to ``catch'' future occurrences.   The analyst can use our visual analysis tool to inspect any cluster toward verifying true vs. false positives (inset \textbf{B}). Notably, this is a critical step as not all statistical outliers are interesting anomalies, due to the ``semantic gap'' \cite{sommer2010outside} (e.g. spikes during Christmas sales). For true/semantic anomalies (i.e. financial fraud), 
we further leverage explanations to
present candidate rules that best capture
the verified anomalous pattern (inset \textbf{C}).
Finally, an interactive interface allows the analyst to revise any of the candidates or design a new rule that can capture these instances (high coverage) but not others (high purity) (inset \textbf{D}).
The newly designed rule(s) are then transferred onto the existing rule database toward flagging similar future anomalies, contributing to supervised detection and thereby closing ``the loop''.
In summary, this work introduces the following main contributions.









\vspace{-0.015in}
\begin{description}[font=$\bullet$\scshape\bfseries,leftmargin=*]
\item \textbf{End-to-end Pipeline for Human-in-the-loop Anomaly Discovery and Management}: We develop a new end-to-end framework, called \method\footnote{\label{note1}\la{\url{https://github.com/xyvivian/ALARM.git}}} (for \underline{\textsc{A}}nalyst-in-the-\underline{\textsc{L}}oop \underline{\textsc{A}}nomaly \underline{\textsc{R}}easoning and \underline{\textsc{M}}anagement), that supports ($i$) unsupervised emerging anomaly detection, ($ii$) human-in-the-loop 
reasoning and verification, and ($iii$) guided action-taking in the form of interactively designing new detection rules for future anomalies of similar nature.

\item \textbf{Anomaly Explanations-by-Design}: 
We equip the unsupervised detection algorithm \ex  
\cite{manzoor2018xstream,sparx22} with model-specific (rather than post hoc/model-agnostic) explanations (i.e. feature importances), capable of handling mixed-type data.
We quantitatively evaluate the accuracy of the feature-importance based explanations by utilizing generative models that simulate mixed-type anomalies.
Notably, explanations are further utilized downstream; for anomalous pattern discovery and candidate rule generation.

\item \textbf{Interactive Visual Toolkit for Verification and Rule Design:} We create a GUI that summarizes detected anomalies in clusters (reducing one-by-one inspection overhead), allows visual inspection and exploration toward verification, and presents candidate rules for interactive, multi-objective rule design  (insets A--D in Fig. \ref{fig:sshot}).

\item \textbf{Financial Application and User Study}:
We employ our end-to-end framework in the financial domain wherein detecting and managing emerging fraudulent schemes in a timely fashion is critical. User studies with three real-world fraud analysts across three case studies and two datasets demonstrate the efficacy and efficiency that \method provides, complementing current practice.

\end{description}

\vspace{-0.015in}
\noindent
{\bf Reproducibility:~} We open-source all code within \method's framework, including front-end GUI and back-end algorithms, publicly.\textsuperscript{\ref{note1}}

\section{Overview \& Background}
\label{sec:prelim}


In our proposed \method pipeline, the first step is effectively detecting the emerging/novel phenomena in the incoming data stream. To this end, we employ one of our own algorithms, called \ex \cite{manzoor2018xstream}, which is recently extended to Apache Spark based distributed anomaly detection \cite{sparx22}. 
\ex is designed for streaming data, and can seamlessly handle feature-evolving, mixed-type data as it appears in many practical applications. Moreover, distributed detection is not only advantageous for real world domains where the data is too large to fit in a single machine, but also when data collection is inherently distributed over many servers, as is the case in the financial bank industry.
Further, our own detection algorithm provides us with full access to the source code, which we extend with built-in (i.e. model-specific) explanations.
Finally, it is efficient and effective; independent studies have found \ex to be very competitive in detecting data stream outliers \cite{ntroumpogiannis2023meta,navarro2022human}.

Two downstream components of our \method framework, namely (1) anomaly explanation and the (2) interactive visual exploration and rule design toolkit, are developed newly as part of the current work in order to assist human analysts in the loop \textit{post} detection, and thus closing the loop from detection to action. 

{\bf Outline.~} Sec. \ref{ssec:xstream} provides a short review of \ex. 
Sec. \ref{sec:explain} presents a model-specific anomaly explanation algorithm for \ex, followed by evaluation of the explanations on simulated mixed-type data with ground truth in Sec. \ref{sec:eval}.
Sec. \ref{sec:toolkit} describes the individual components of our \method toolkit and Sec. \ref{sec:ustudy} presents the user study results.
We conclude the paper in Sec. \ref{sec:conclusion}. 

\vspace{-0.1in}
\subsection{Anomaly Detection with \ex: Review}
\label{ssec:xstream}

\ex 
 consists of three main steps, which we review briefly for the paper to be self-contained, and refer to \cite{manzoor2018xstream} 
 for details.

\vspace{-0.05in}
\subsubsection{\bf Step 1. Data Projection}
\label{sssec:s1}

Given mixed-type data $\bx_i \in \R^D$, 
\ex creates a low-dim. sketch $\bs_i$ 
via random projections \cite{indyk1998approximate,achlioptas2003database}: 
\beq
\label{eq:sketch}
\bs_i = (\bx_i^T \br_1, \dots, \bx_i^T\br_K)
\eeq
where $\{\br_1,\ldots,\br_K\}$
depict $K$ 
\textit{sparse} random vectors s.t.  $\br_k[F]\in \{\pm1\}$  with prob. 1/3, and zero otherwise \cite{achlioptas2003database}.  
For streaming data, entries of $\br_k$ is computed \textit{on-the-fly} via hashing, rewriting Eq. \eqref{eq:sketch} as
\beq
\label{eq:hash}
\scalemath{0.85}{
\bs_i[k] =  \sum_{F \in \mF_r} h_k(F) \;\cdot\; \bx_i[F]  
\;+ \sum_{F \in \mF_c}  h_k(F \oplus \bx_i[F]) \;\cdot\; 1
\;,\;\; k=1\ldots K
}
\eeq
\vspace{-0.1in}

\noindent
where $h_k(\cdot)$ is a hash function, $\mF_r$ and $\mF_c$ respectively denote the set of real-valued and categorical features, $\bx_i[F]$ is point $i$'s value of feature $F$, and $\oplus$ depicts the string-concatenation. 

\hide{
When triplet updates $<ID, F, \delta>$ arrive over the stream, where $\delta =\;$\texttt{\small{old\_val:new\_val}} for categorical features, the sketch can be updated by
\begin{equation}
\label{eq:update}
\hspace{-0.125in}
\scalemath{0.9}{
\bs_{ID}[k] = \begin{cases}
 \bs_{ID}[k] + h_k(F) \cdot \delta   {\text{\quad if real-valued $F$,}}
\\
\bs_{ID}[k] - h_k(F\oplus\text{\texttt{\small{old\_val}}}) + h_k(F\oplus\text{\texttt{\small{new\_val}}})   \text{  o.w.} 
\end{cases}
}
\end{equation}
for $k=1\ldots K$ such that $h_k(F\oplus\text{\texttt{\small{old\_val}}})$ returns zero when \texttt{\small{old\_val}} is \texttt{\small{null}}.
It is important to notice that Eq. \eqref{eq:update} can seamlessly handle a newly emerging feature $F$ that has never been seen before.
}

\vspace{-0.05in}
\subsubsection{\bf Step 2. Denstiy Estimation with Half-space Chains}
\label{sssec:s2}

Anomaly detection relies on density estimation at multiple scales via a set of so-called Half-space Chains (HC), a data structure akin to multi-granular subspace histograms.
Each HC has a length $L$, along which the (projected) feature space $\mF_p$ is recursively halved on a randomly sampled (with replacement) feature, where $f_l \in \{1,\ldots,K\}$ denotes the feature at level $l=1,\ldots, L$.
In general, a point can lie in one of $2^l$ bins at level $l$.

Given a sketch $\bs$, the goal is to efficiently identify the bin it falls into at each level.
Let $\bdel \in \R^K$ be the vector of initial bin widths, equal to half
the range of the projected data along each dimension $f \in \mF_p$.
Let $\bzb_l \in \Z^K$ denote the bin identifier of $\bs$ at level $l$, initially all zeros.
At level 1, bin-id is updated as $\bzb_1[f_1] = \lfloor \bs[f_1]/\bdel[f_1] \rfloor$.
At consecutive levels, it  can be computed \textit{incrementally}, as

\vspace{-0.15in}
\begin{equation}
\label{eq:binid}
\scalemath{0.9}{
\bzb_l[f_l] = \lfloor \bz_l[f_l] \rfloor \;\text{  s.t.  } \;
\bz_l[f_l] = \begin{cases}
\bs[f_l]/\bdel[f_l]  \text{\quad if $o(f_l,l) =1$, and}
\\
2\bz_l[f_l] \text{\quad \quad\;\; o.w.; if $o(f_l,l)> 1$ } 
\end{cases}
}
\end{equation}
where $o(f_l,l)$ denotes the number
of times feature $f_l = \{1,\ldots,K\}$ has been sampled in the chain until and
including level $l$.

Notice that all points with the same unique $\bzb_l$ lie in the same histogram bin at level $l$. Then, level-wise (multi-scale) densities are estimated by counting the number of points with the same bin-id per level. 
\ex obtains approximate counts via a count-min-sketch \cite{cormode2005improved}, the size of which is user-specified, i.e. constant.

Overall, \ex is an ensemble of $M$ HCs, $\mH = \{HC^{(m)} := (\bdel, \bfl^{(m)}, \mC^{(m)})\}_{m=1}^M$ where each HC is associated with 
 (i) bin-width per feature $\bdel \in \R^K$,
 (ii) sampled feature per level $\bfl^{(m)} \in \Z^L$,
and (iii) counting data structure per level $\mC^{(m)} = \{C_l^{(m)}\}_{l=1}^L$.

\vspace{-0.05in}
\subsubsection{\bf Step 3. Anomaly Scoring}
\label{sssec:s3}
To score a point for anomalousness, count of points in the bin that its sketch falls into at each level $l$ of a HC, denoted $C_l^{(HC)}[\bzb_l]$,
 is extrapolated via multiplying by $2^l$ s.t. the counts are comparable across levels. 
 Then, the smallest extrapolated count 
 is considered the anomaly score, i.e. 

\vspace{-0.15in}
\beq
\label{eq:score}
\scalemath{0.9}{
O^{(m)}(\bs) = 
\min_l \; 2^l \cdot C_l^{(m)}[\bzb_l] \;.
}
\eeq
The average across all HCs is the final anomaly score; the lower the score, the lower is the density and higher the anomalousness. 


\section{Anomaly Explanation}
\label{sec:explain}


Given the detected anomalies by \ex, we aim for \textit{model-specific} explanations per anomaly, i.e., individual explanations. 
As detection is based on density estimates in feature subspaces, explanations aim to reflect feature importances. Specifically,

\noindent
\textbf{Given} (1) a trained set of half-space chains $\mH = HC^{(1)}, \ldots, HC^{(M)}$, and (2) a detected anomaly point $\bx$;

\noindent
\textbf{Estimate} importance weights for the original features $\mF_r \cup \mF_c$.

\vspace{-0.1in}
\subsection{Estimating Feature Importances}
\label{ssec:fimp}

To estimate the weight of a feature for a high-score anomaly, we follow a simple procedure that leverages the ensemble nature of \ex. 
In a nutshell, it identifies the half-space chains in the ensemble that ``use'' the feature in binning the feature space, 
and (re)calculates the 
the anomaly score of the point only based on this set of chains. The higher it is, 
the more important the feature is  in assigning a high score to the (anomalous) point.

Specifically, recall from Sec. \ref{sssec:s2}
that $\bfl^{(m)} \in \Z^L$ denotes the sequence of features used in halving the feature space by chain $m$. Given $M$ chains $\mH = \{HC^{(m)}\}_{m=1}^M$, and a feature $f$ to estimate its importance for a (projected) point $\bs$, we partition the chains into two groups: those that do and do not ``use'' $f$ in $\bfl^{(m)}$. 

The definition of ``use'' needs care here, due to how the anomaly score of a point  is estimated by a chain. Note in Eq. \eqref{eq:score} that the level $l$ at which the extrapolated count is the minimum provides the score; in effect, only the features up to $l$ contribute to a point's score. 
Therefore, a feature is considered ``used'' by a chain if it is a halving feature from the top down to this scoring level only.

Let $l_{\bs}$ denote the level at which a point $\bs$ is scored by a chain. 
A feature $f$ is used by chain $m$ if 
$f \in \bfl^{(m)}[1]\ldots \bfl^{(m)}[l_\bs]$.  
Let $\mathcal{M}_u^{(f)}$ 
denote the chain indices that use 
feature $f$. Then, the importance weight of $f$ for point $\bs$ is given as

\vspace{-0.1in}
\beq
\label{eq:importance}
\scalemath{0.9}{
w(f\vert\bs) = \frac{1}{|\mathcal{M}_u^{(f)}|} \sum_{m \in \mathcal{M}_u^{(f)}} 
O^{(m)}(\bs)
\;.
}
\eeq
\vspace{-0.05in}

Note that feature importances 
differ by point, and hence are individualized, since $l_{\bs}$ is dependent on the input point. 

We note that several alternative importance measures did not perform well, such as the {difference} between scores from the chains that do and do not use $f$, or the drop in the anomaly score when the chains that use $f$ are removed. The reason is  multicollinearity; 
when chains that did not use an important feature $f$ used correlated features instead, they continued to yield a high anomaly score.

\vspace{-0.1in}
\subsection{From Projected to Original Features}
\label{ssec:orig}

Recall from Sec. \ref{sssec:s1} that \ex creates projection features $k=\{1\ldots K\}$ to sketch high-dimensional and/or mixed-type data.
The chains are built using the projected features, thus, the estimated importances above are for those ``compound'' features.

Consider a projection feature $k$ with the corresponding sparse hash function $h_k(\cdot)$, which outputs 
e.g. $h_k($`$Gender\|Female$'$)=-1$,
$h_k($`$Age$'$)=1$, and $0$ o.w. 
For a Female with (normalized) Age 0.6 (and possibly other features), it takes the value $-0.4$. Then, the importance of a projection feature needs to be ``attributed'' back to the original features in its compound, in this case Gender and Age.

To this end, the relations between the projected and original features can be captured as a sparse bipartite graph.  Nodes $k=\{1\ldots K\}$ on one side depict the projected features with pre-computed feature importances (node weights)  as described in Sec. \ref{ssec:fimp}.
Nodes $F=\{1\ldots |\mF_r \cup \mF_c|\}$ on the other side depict the original features. 
Note that this graph is built separately for each (anomalous) point $\bx$ to be explained.
Thanks to the binary hash functions, there exists an edge $(k,F)$ only when $h_k(F) \neq 0$ (or for categorical F, when $h_k(F \oplus \bx[F])\neq 0$), with expected density 1/3.

To attribute importances from projection features to the original features, a simple approach could sum the importances of the projection features whose compound an original feature participates in (i.e. sum of neighbors' weights). However, this may attribute  spurious importance from a neighbor that is important due to a different feature in its compound. 
It would also fail to tease apart additive feature attributions \cite{lundberg2017unified}
in the presence of multicollinearity.
Admittedly, estimating individual direct-effect importances would be combinatorially hard. As an intermediate solution, we go beyond the direct neighbors and 
diffuse in the graph the initial projection feature weights via random walk with restart \cite{haveliwala2002topic}.

Let $\bA \in \{0,1\}^{(K\times |\mF_r \cup \mF_c|)}$ denote the  adjacency matrix of the bipartite graph,
$\bpi = \bpi_o \| \bpi_p$ denote the concatenated `topic-sensitive' Pagerank vector for the \textit{\textbf{o}}riginal and \textit{\textbf{p}}rojected features, respectively, and 
$\bw_p$ be 
the vector of projected feature importance weights based on Eq. \eqref{eq:importance}.
$\bw_p$ is normalized to capture the fly-back probabilities, and $\bpi$ is initialized randomly and normalized over iterations. Then,

\vspace{-0.1in}
{\small{
\begin{align}
    \bpi_p^{(t+1)} & := (1-\alpha) \times \bA \bpi_o^{(t)} + \alpha \times \bw_p \\
    \bpi_o^{(t+1)} & := (1-\alpha) \times \bA^T \bpi_p^{(t+1)}
\end{align}
}}

\noindent
iteratively compute the original feature importances $\bpi_o$, using the restart probability $\alpha=0.15$.

\section{Evaluation: Anomaly Explanation}
\label{sec:eval}
\subsection{Simulation}
\label{ssec:sim}

To assess the performance of our feature explanations, we require data containing anomalies with ground-truth feature importance weights, which (to our knowledge) does not publicly exist. 
To this end, we create a new simulator synthesizing anomalies in subspaces along with feature importances. We publish the simulator source code \ref{note1}, which may be of independent interest to XAI communities. 


A basic simulator could generate data from a predefined distribution, altering subset of features to create anomalies. However, it may produce unrealistic data, falling short in determining the importance of altered features. Consider the case where features A and B are expanded by different factors (5 times and 10 times respectively). It is not clear which feature, A or B, is more responsible for outlierness. This lack of clarity in determining feature importance makes it difficult to establish a definitive ground truth.

To overcome the above difficulties, we propose using generative models with real-world data to synthesize anomalous points and feature importances. 
We use the variational auto-encoder (VAE) \cite{kingma2013vae} 
to capture complex data distributions with both real-valued and categorical features. VAE contains an encoder-decoder couple, parameterized by $\phi$ and $\theta$ respectively. It embeds a training point $\textbf{x}$ into a lower-dim. $\textbf{z}$. At training stage, the encoder minimizes the distance of a surrogate posterior $q_\phi (\textbf{x} | \textbf{z})$ to the true  $p_\theta (\textbf{z})$,
while the decoder maximizes $p_\theta (\textbf{x} | \textbf{z}) $, the probability of $\textbf{x}$ 
given $\textbf{z}$. The likelihood of a point being an anomaly can be determined by VAE's reconstruction probability $p_\theta (\textbf{x} | \textbf{z})$. 
We can also use $p_\theta (\textbf{x} | \textbf{z})$ to calculate feature importances, by comparing the reconstruction probability when a feature is altered to that when it is not.

Algo. \ref{alg:cap} lists the steps of our simulator. 
Given a dataset, we feed all the normal points into the VAE and generate $m$ normal points $\{\textbf{x}_{nor}^i\}_{i=1}^{m}$ with $\{\textbf{z}^i\}_{i=1}^{m}$ hidden variables  (lines 1-2), using the Gaussian prior $\mathcal{N}(0,1)$ (although VAEs can be customized to better fit the data distribution using priors like GMM \cite{dilokhanakuldeep2016}, Gumbel \cite{jang2016gumbel}, etc.). Then, we set a threshold $\tau$ for specifying anomalous points as

\vspace{-0.15in}
\begin{align}
\tau = \epsilon \cdot  \min_{i=1,...,m}[ \log p_{\theta} (\textbf{x}_{nor}^i | \textbf{z}^i) ] \;,
\end{align}
\vspace{-0.15in}

\noindent
 which is the scaled minimum likelihood of a point being normal, 
  where scale $\epsilon$ specifies the tightness of the threshold (line 3). 

Next we generate the anomalies and their associated feature importance vectors (lines 6-13). 
For each anomaly, we first sample a normal instance $\textbf{x}$ with hidden $\textbf{z}$. 
Given a subset of features ``to-inflate'', which can differ per anomaly, we ``inflate''  point $\textbf{x}$ along each specific dimension $j$ to acquire $\textbf{x}^{\text{inflated}j}$. For real-valued $j$, the point is placed in a low-density region far from normal points (global) or in the vicinity of normal points yet with low probability (local). For categorical $j$, the point's value is replaced by one with lower probability, calculated from the empirical distribution (See Appx. \ref{sssec:inflate}).
Then, 
feature $j$'s  importance weight is calculated as

\vspace{-0.15in}
\begin{align}
\label{eq:fiw}
\textbf{e}_j = \max \{ \; 0, \; \log p_{\theta} (\textbf{x} | \textbf{z}) - \log p_{\theta} (\textbf{x}^{\text{inflated}j} | \textbf{z}) \; \} \;.
\end{align}
\vspace{-0.15in}

\noindent


If the posterior log-probability $\log p_{\theta} (\textbf{x}_{\text{candidate}} | \textbf{z})$ (a.k.a. anomaly score $s$) of 
point $\textbf{x}_\text{candidate}$, 
 with inflated values for all features in the subset, 
 is low, i.e. smaller than threshold $\tau$, then it is added to the anomaly pool along with its feature importances 
$\textbf{e}$ (lines 14-18). We continue this process until $k$ anomalies are generated.

{\small{
 \begin{algorithm}[th]
\caption{\textbf{Anomaly \& Feature Importance  Synthesis} 
}\label{alg:cap}
\begin{algorithmic}[1]
\Require $d$-dim. normal data $\{ \textbf{x}_{train}\}$, $\;m$: num. synth. normal pts, $\;k$: num. synth. anomalies, $\epsilon$: scalar,  feature indices to-inflate

\State Train a VAE model $VAE(\phi, \theta)$ with $\{ \textbf{x}_{train}\}$ \label{lst:line:1}

\State Sample $\{\textbf{x}_{nor}^i\}_{i=1}^{m} \sim VAE(\phi, \theta)$ with $\{\textbf{z}^i\}_{i=1}^{m}$ \label{lst:line:2}
\State Set anomaly threshold $\tau := \epsilon \cdot  \min_{i=1,...,m}[ \log p_{\theta}(\textbf{x}_{nor}^i | \textbf{z}^i) ]$.
\label{lst:line:3}
\State Initialize empty {\sf AnomalyList} and {\sf FeatureImportanceList}
\label{lst:line:4}

\While{length({\sf AnomalyList}) $< k$} \label{lst:line:5}
    \State Sample  $\textbf{x} \sim VAE(\phi,\theta) $with $\textbf{z}$ \label{lst:line:6}
    \State Initialize  $d$-dim. vector $\textbf{e}$ with zeros \label{lst:line:7}
    \State Let $\textbf{x}_\text{candidate} \gets \text{copy}(\textbf{x})$ \label{lst:line:8}
    \For{(each to-inflate feature $j \in [1,...,d] $)} \label{lst:line:9}
        \State Inflate feature $j$ of $\textbf{x}$ to obtain $\textbf{x}^{\text{inflated}j}$ \label{lst:line:10}
         \State Let $\textbf{e}_j \gets \max \{ 0, \log p_{\theta} (\textbf{x} | \textbf{z}) - \log p_{\theta} (\textbf{x}^{\text{inflated}j} | \textbf{z}) \}$
         \label{lst:line:11}
         \State Let $\textbf{x}_{\text{candidate},j} \gets \textbf{x}^{\text{inflated}j}_j$ \label{lst:line:12}
    \EndFor \label{lst:line:13}
    \State  Compute $s \gets \log p_{\theta} (\textbf{x}_{\text{candidate}} | \textbf{z}) $
    \label{lst:line:14}
    \If{ $s < \tau$}   \label{lst:line:15}
     \State
     {\sf AnomalyList}.append($\textbf{x}_\text{candidate}$) \label{lst:line:17}
        \State {\sf FeatureImportanceList}.append($\textbf{e}$)   \label{lst:line:18}
\EndIf \label{lst:line:19}
\EndWhile \label{lst:line:20}

\State {\bfseries Return:} $\{\textbf{x}_{nor}^i\}_{i=1}^{m}$, {\sf AnomalyList}, {\sf FeatureImportanceList} \label{lst:line:21}
\end{algorithmic}
\end{algorithm}
}}
\setlength{\textfloatsep}{0.1in}

\subsection{Experiment Setup} 

{\bf Data:~} 
We evaluate our anomaly explanation-by-feature importances approach on three real-valued and three mixed-type datasets, which are commonly used in anomaly detection literature for tabular data.
Table \ref{tab: datasetoverview} lists the dataset names and descriptions. {All data are publicly available at the UCI machine learning repository \cite{Dua2017UCI}.\footnote{Also downloadable from \url{http://odds.cs.stonybrook.edu}}}


\begin{wraptable}{r}{3.75cm}
\vspace{-0.15in}
\centering
\caption{Dataset statistics. }
\vspace{-0.15in}
	\centering
	\hspace{-0.25in}
	{\scalebox{0.8}{
    \begin{tabular}{lrrrr}
    \toprule
    \textbf{Name} &  \textbf{Type} &
    $|\mF_c|$
    &   
    $|\mF_r|$ \\
    \midrule
    Seismic & Mixed &  4 & 11 \\
    KDDCUP & Mixed &  3 & 31 \\
    Hypothyroid & Mixed &  12 & 6 \\
    Cardio & Real-val & 0 & 21 \\
    Satellite & Real-val & 0 & 36 \\
    BreastW & Real-val & 0 & 9 \\
    \bottomrule
    \end{tabular}
}}
\vspace{-0.1in}
    \label{tab: datasetoverview}
\end{wraptable} 
For each dataset, we pre-process the data by removing any points with missing features and any features that only have one value. We then use all the cleaned normal points to fit a VAE and generate 5000 normal points and 500 anomalies. {To generate an anomaly, we inflate 1/3 of the dataset's features, randomly chosen for each anomaly. Real-valued features are inflated to yield at random either global or local anomalies, as described in Sec. \ref{ssec:sim}. Categorical features are inflated by replacing the original value with that of  lowest probability.} 
{Associated feature importance weights are obtained based on our anomaly simulator, specifically Eq. \eqref{eq:fiw},  and used as ground-truth for evaluating the explanations.}


\noindent 
{\bf Baselines:~} 
We evaluate our method along with two popular explanation methods: SHapley Additive exPlanations (SHAP) 
 \cite{lundberg2017unified} and Depth-based Isolation Forest Feature Importance (DIFFI) \cite{diffi2022carletti}.
SHAP is a method for interpreting the output of any machine learning model that is based on Shapley values and is model-free. SHAP's feature importance can be computed using the predictions of both \ex as well as Isolation Forest (IF)
\cite{liu2008isolation} algorithm---one of the state-of-the-art anomaly detection methods for tabular data \cite{emmott2015meta}. In contrast, DIFFI is a feature importance method that is specifically based on using IF as the backbone anomaly detection method. 

In addition, we compare feature importances by \ex without  as well as with feature projection to varying dimensions for $K$. 
We set the other hyperparameter values sufficiently large as suggested in \cite{manzoor2018xstream} so as to obtain good detection performance. 
Specific configurations and other setup details can be found in Appx. \ref{ssec:expeval}.


\begin{table*}[!th]
\caption{Ranking quality of features by importance produced by \ex and baselines, measured by NDCG.  ``$\dagger$'' indicates cases when detection performance is larger than 0.99 AUROC. In parenthesis is the recorded average time to acquire the explanation. }
\vspace{-0.15in}
\begin{tabular}{r c c c c c c}
\hline
{Method} & {Seismic} & {KDDCUP} & {Hypothyroid} & {Cardio} & {Satellite} & {BreastW} \\
\hline
IF+SHAP ($\sim 5$ mins) & 
\textbf{0.874$\pm$0.002}$^{\dagger}$ &
\textbf{0.740$\pm$0.010}$^{\dagger}$ & 
0.810$\pm$0.011  & 
0.828$\pm$0.005$^{\dagger}$ & 
0.907$\pm$0.001$^{\dagger}$ &
\textbf{0.957$\pm$0.008}$^{\dagger}$ \\

DIFFI ($\leq 1$ min) &
0.867$\pm$0.013$^{\dagger}$  & 
0.726$\pm$0.008$^{\dagger}$ & 
0.824$\pm$0.023  & 
0.823$\pm$0.004$^{\dagger}$ & 
0.868$\pm$0.005$^{\dagger}$ & 
0.912$\pm$0.006$^{\dagger}$  \\

\ex+SHAP ($\geq 130$ hrs)&
 \textbf{0.875$\pm$0.006}$^{\dagger}$&
 0.695$\pm$0.007 &
 \textbf{0.830$\pm$0.010}$^{\dagger}$&
\textbf{0.847$\pm$0.005}$^{\dagger}$ &
\textbf{0.912$\pm$0.003}$^{\dagger}$&
\textbf{0.955$\pm$0.007}$^{\dagger}$ \\
\ex \textbf{w/out} proj. ($\leq 1$ min)& 
\textbf{0.873$\pm$0.018}$^{\dagger}$ &
0.670$\pm$0.010 & 
\textbf{0.828$\pm$0.020}$^{\dagger}$& 
0.836$\pm$0.008$^{\dagger}$&
\textbf{0.910$\pm$0.006}$^{\dagger}$&
0.827$\pm$0.005$^{\dagger}$\\
\hline
\end{tabular}
\label{tab:result}
\vspace{-0.05in}
\end{table*}

\begin{table*}[h]
\caption{Ranking quality of features by importance using \ex with projection; using varying number of projection dimensions $K$, measured by NDCG. ``$*$'' and ``$\dagger$'' indicate cases when detection performance is larger than 0.95 and 0.99 AUROC, respectively. In parenthesis is the recorded average time to acquire the explanation.}
\vspace{-0.15in}
\begin{tabular}{c c c c c c c}
\hline
{Method} & {Seismic} & {KDDCUP} & {Hypothyroid} & {Cardio} & {Satellite} & {BreastW} \\
\hline
\ex \textbf{w/} proj. $K$=15 ($\leq 1$ min) &
 0.666$\pm$0.006 $^*$&
 0.424$\pm$0.006 &
 0.532$\pm$ 0.024&
 0.706$\pm$0.005 $^{\dagger}$& 
0.628$\pm$0.006 $^{\dagger}$&
0.837$\pm$0.004 $^{\dagger}$\\
\ex \textbf{w/} proj. $K$=20 ($\leq 1$ min)&
0.688$\pm$0.002 $^*$&
0.444$\pm$0.013&
0.560$\pm$0.008&
0.713$\pm$0.009 $^{\dagger}$&
0.652$\pm$0.004 $^{\dagger}$&
0.850$\pm$0.005 $^{\dagger}$\\ 
\ex \textbf{w/} proj. $K$=30 ($\leq 1$ min) &
0.702$\pm$0.003 $^*$&
0.488$\pm$0.006 $^*$&
0.542$\pm$0.016 &
0.752$\pm$0.005 $^{\dagger}$&
0.675$\pm$0.007 $^{\dagger}$&
0.865$\pm$0.004 $^{\dagger}$\\
\hline
\end{tabular}
\label{tab:proj_result}
\vspace{-0.1in}
\end{table*}

\noindent 
{\bf Metrics:~} 
The main metric for evaluation is ranking based, quantifying how well we rank the features by importance; namely 
Normalized Discounted Cumulative Gain (NDCG) \cite{ndcg2000}. NDCG compares the effectiveness of a ranking to an ideal ranking, summing the relevance-weighted scores of the items in the predicted ranking. 
We prefer NDCG as it i) gives more weight to the top anomalous features (we apply $\log 2$ as the base of the discount factor), and ii) can use ground-truth feature importances as the relevance score. 

We also measure the time it takes for \ex and other comparison methods to acquire feature importance explanations, as well as the performances of the underlying detection methods, measured by Area Under the Receiver Operating Characteristic (AUROC).

\vspace{-0.15in}
\subsection{Results}
Table \ref{tab:result} displays the NDCG ranking quality w.r.t. to the synthesized ground-truth feature importances, as well as the approximate computational time required,  comparing \ex (without projection) and various baseline methods. The highest NDCG scores are achieved with the combination \ex+SHAP (for detection+explanation, respectively) for almost all datasets. However, it is computationally quite demanding, taking more than 130 hours. The  IF+SHAP combination provides a faster solution, delivering results in about 5 minutes, as it uses sped-up computations of SHAP \cite{lundberg2020local} for tree-based methods like IF. \ex and DIFFI, two model-specific explanation methods, are even faster. \ex (w/out projection) is comparable to state-of-the-art explanation models or often the runner-up for many of the datasets. Importantly, \ex can be applied to distributed and/or streaming data, which makes it more appealing and practical for large data real-world systems, in comparison to DIFFI and IF+SHAP.

Table \ref{tab:proj_result} shows the NDCG scores of \ex with different number of projections. The usage of projection diminishes \ex's capability to detect and subsequently explain the anomalies. In other words, when projection is used there is a noticeable decrease in both AUROC and NDCG. The decline may be driven by two factors. First, when \ex is used with projection, its detection accuracy decreases which associates with lesser quality chains, making it difficult to obtain an explanation (See exact AUROC detection performances in Table \ref{tab:auroc} in Appx. \ref{ssec:expeval}). Second, the graph propagation-based attribution is a heuristic and may not be accurate in fully capturing the direct feature effects. Nevertheless, the use of projection allows \ex to explain feature-evolving streaming data without requiring a complete retraining of the algorithm, making it more suitable for real-time applications.

Besides quantitative comparison, we also note disagreement among the explanations themselves, 
where different methods 
yield feature importances with significant variations \cite{disagreement2022hima}. This suggests that 
no explanation 
can be considered the definitive truth for end-users, and highlights the importance of the human in the loop: rather than blindly accepting the feature importances produced by any specific algorithm, analysts should be able to actively participate in the anomaly mining, reasoning, and management cycle.

\vspace{0.1in}
\section{From Explanation to Action: A New Toolkit for Anomaly Management}
\label{sec:toolkit}

The premise of
anomaly explanations is to equip the human-in-the-loop with a deeper insight and understanding regarding the nature of the flagged anomalies. However, explanations are only as valuable as they are \textit{useful} for the analysts \cite{kaur2020interpreting}, ideally in improving a downstream task with a measurable objective \cite{shen2020useful,jesus2021can}.

In many real world scenarios, including our financial application domain, the analyst's main goal is to derive enough knowledge from the explanations so as to be able to \textit{prevent future anomalies} of the same nature. The action toward that goal may be fixing or troubleshooting various components of a system that the analyst has access to and full control over. In other, especially adversarial scenarios, the action may involve instigating new policies regarding how the system is allowed to operate in the future.

Particularly in the financial domain, among others, the analyst aims to deploy a \textit{new detection rule} for the potential recurrences of the detected threat. 
Rule-based detection systems are typical of many deployed applications in the real world for several reasons. First, rules are simple; they are short and readable by humans. Second, they enable fast filtering of potentially streaming incoming data. Moreover, a database or ensemble of rules allow multiple analysts to populate the database with new rules independently, in a decentralized fashion. 
Therefore, our overarching approach to putting explanations into action is to build a new toolkit that facilitates designing new rules for emerging threats. The toolkit is to allow inspecting and attending not only to the anomalies as detected by an algorithm but also to those as reported by external sources (e.g. other banks, card customers, etc.). 

To best support the human in the loop, we build a 
\textit{visual} and \textit{interactive} graphical user interface (GUI) for \method. 
It consists of four main building blocks, as detailed in Sections \ref{ssec:clus}--\ref{ssec:rdi}, that respectively fulfill four key design requirements.

\sloppy{
\textbf{First}, the anomalies need to be summarized---by grouping similar anomalies---as individually inspecting each anomaly would be too time-consuming in presence of several hundred that are flagged.
Further, analysts are interested to capture anomalous groups or patterns, indicative of repeating anomalies (that may continue to emerge in the future), rather than one-off anomalies.
\textbf{Second}, human analysts often prefer visually inspecting how the data generally look like and how the anomalies stand out. Their main goal in inspecting is to verify if the anomalies are truly semantically relevant or otherwise false positives. 
}
\textbf{Third}, analysts could benefit from automatically generated candidate rules. Data-driven rules provide a reasonable starting point that the analyst can revise, reducing the time from detection to response. Importantly, some analysts may be more novice than others and find a starting point helpful.
\textbf{Finally}, the GUI should support fully interactive rule design that allows adding/removing feature predicates. Analysts often have years of expertise in identifying useful predicates that capture recurring attack vectors. They may also prefer some features over others for their cost-efficiency (easier or faster to track) as well as for various policy reasons (justifiable, privacy-preserving, ethical, etc.).

The following presents the four building blocks of \method's GUI component, implementing the wish-list above.

\begin{figure}[h]
	\centering
\includegraphics[width=0.9\linewidth]{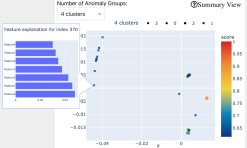}
\vspace{-0.1in}
	\caption{Summary View groups the   anomalies into user-given \# clusters (membership by symbol), and also conveys the score (color) and feature importances (per anomaly). \label{fig:summary}}
		\vspace{-0.15in}
\end{figure}

\vspace{-0.1in}
\subsection{Summary View (\clus):~}
\label{ssec:clus}

Given a list of anomalies to be inspected, the summary view clusters them by similarity. Similarity is based on the feature importances as estimated by anomaly explanation (Sec. \ref{sec:explain}), rather than feature values in the original space. That is, we use $\text{sim}(\bpi_{o,i}, \bpi_{o,i'})$ that helps group the points that stand out as anomalous in similar subspaces.

\begin{figure*}[!th]
	\centering
\includegraphics[width=1.00\linewidth]{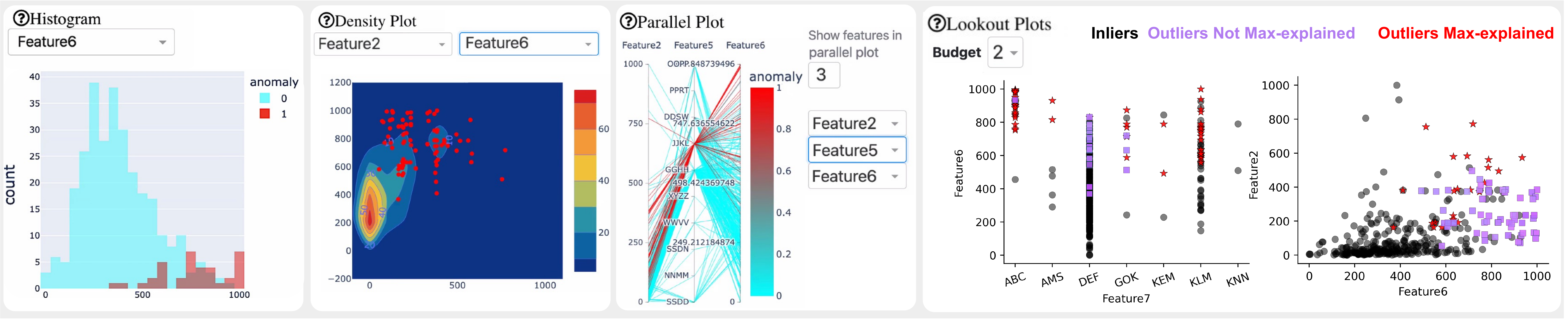}
\vspace{-0.25in}
	\caption{Exploration View offers visual tools that allow inspection of the anomalies, as contrasted with the non-anomalies.\label{fig:explore}}
		\vspace{-0.125in}
\end{figure*}

As Fig. \ref{fig:summary} illustrates, anomalies are presented in a 2-dimensional MDS embedding space \cite{kruskal1978multidimensional} for visualization, which preserves the aforementioned pairwise similarities as best as possible. The analyst can choose the number of clusters, where different symbols depict the cluster membership of the anomalies. The color of the points reflect their anomaly score from \ex. 
Hovering over or clicking an individual point opens up a window that shows feature importances with horizontal bars.

We remark that while the anomalies to be inspected could be the top $k$ highest scoring points from \ex, \method also allows the analyst to import points of their own interest to inspect; e.g. (labeled) anomalies obtained via external reporting. In that case,  the analyst data is passed onto \ex (ignoring the labels), which provide anomaly scores and explanations for the labeled points. Summary View displays only these labeled anomalies of interest, with explanations from \ex that are used toward clustering.

\vspace{-0.1in}
\subsection{Exploration View (\expl):~}
\label{ssec:expl}

Fig. \ref{fig:explore} illustrates the tools that \method offers for data exploration, consisting of four main components that can aid {sense-making and verification}. First, \textbf{Histogram} displays the distributions of anomalous and inlier points along a single dimension. Second, \textbf{Density plot} shows the scatter of anomalous points (red dots) and the density of inliers (heatmap) along two feature dimensions. Third, \textbf{Parallel plot} displays the comparison of anomalies vs. inliers in a multi-dimensional setting, where each is represented by a separate polyline. For all three components, the analyst is able to select which features to display. Finally, \textbf{Lookout} \cite{gupta2019beyond} presents a few scatter plots which are automatically selected feature pairs that maximally-explain (``maxplain'') the anomalies in two dimensions, i.e. wherein  the anomalies stand out the most.
 Analyst can choose the budget interactively, adjusting the number of plots that can be used to maxplain all the anomalies.

\vspace{-0.1in}
\subsection{Rule Candidates View (\cand):~}
\label{ssec:cand}

Given a group of anomalies, x-PACS algorithm \cite{macha2018explaining} generates concise rules, with a small set of predicates, that characterize the anomalous pattern. 
It estimates the univariate kernel and histogram density of the anomalous points, respectively for each numerical and nominal feature, to identify the intervals or values of significant peaks. It then combines these from selected features, where the feature intervals that define the peaks are presented as the predicates. 

\begin{figure}
\begin{minipage}[c]{0.47\linewidth}
\includegraphics[width=\linewidth]{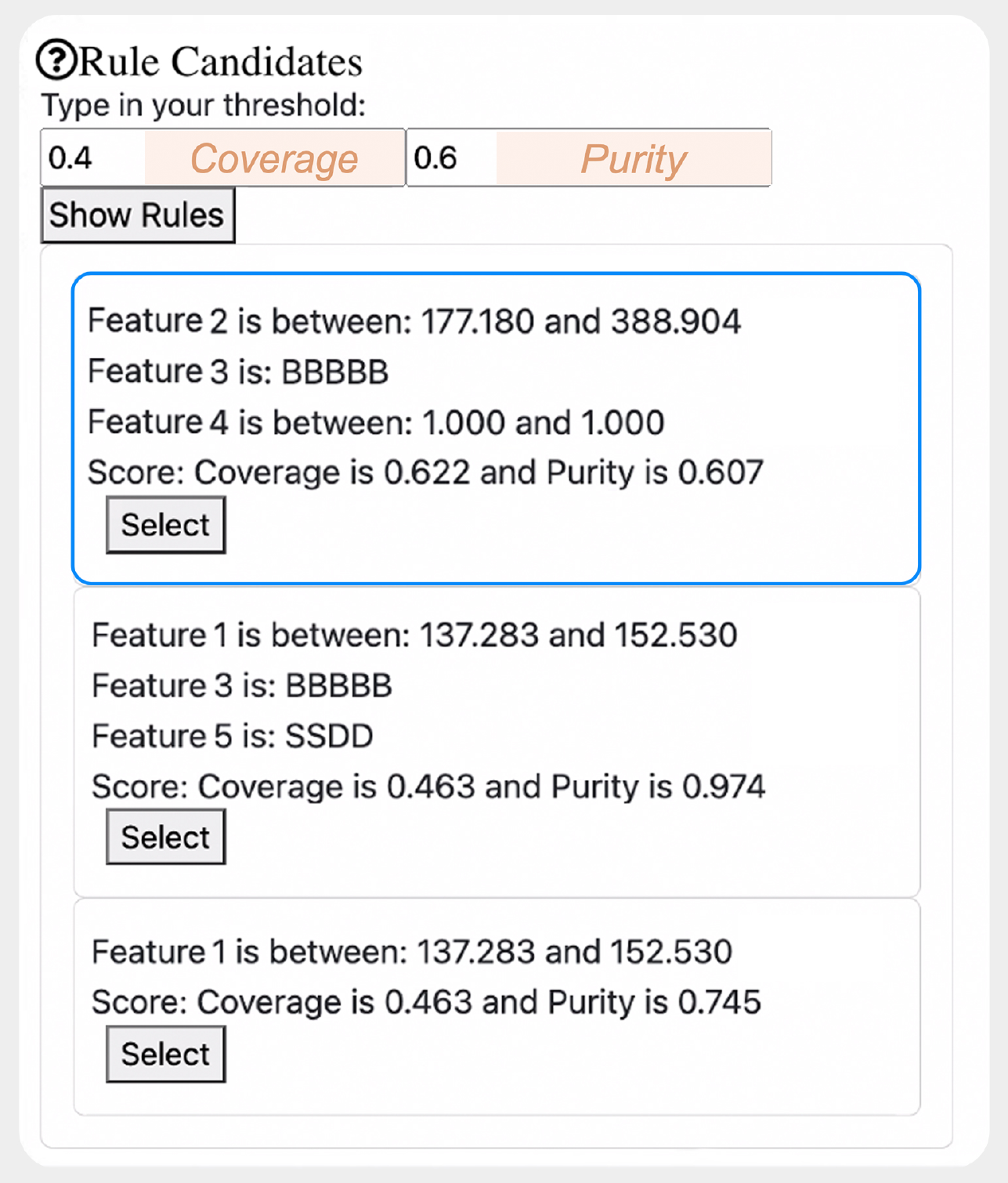}
\vspace{-0.25in}
\caption{Rule Candidates View displays up to three data-driven rules that satisfy typed-in thresholds.\label{fig:rule_candidates}}
\end{minipage}
\hfill
\begin{minipage}[c]{0.51\linewidth}
\includegraphics[width=\linewidth]{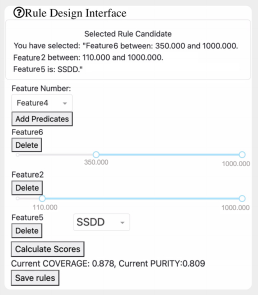}
\vspace{-0.25in}
\caption{Rule Design Interface allows fully-interactive rule generation along with associated metrics.\label{fig:rule_designs}}
\end{minipage}%
\end{figure}

Fig. \ref{fig:rule_candidates} shows a screenshot of our Rule Candidates View, which displays up to three rules that satisfy two user-specified thresholds: coverage (C) and purity (P). C is
the fraction of anomalies in the group that comply with the rule, and  
P is the fraction of inlier points that do \textit{not} pass the rule. 
As such, the higher both C and P are, the better, as they associate with high recall but low false alarm rates. 

\vspace{-0.1in}
\subsection{Rule Design Interface (\rdi):~}
\label{ssec:rdi}

Fig. \ref{fig:rule_designs} illustrates a screenshot of \method's interface  
toward facilitating the design of a new rule with high coverage and purity. The analyst can select a candidate rule to revise or design a rule from scratch by adding or deleting features, and adjusting their values.  
Real-valued feature predicates are adjusted by sliders that allow specifying intervals, and categorical features can be assigned a value by scrolling through a drop-down list.  ``Calculate Scores'' button displays the coverage and purity of the latest set of predicates. Upon completion, the rule can be saved locally or in a rule database used for supervised anomaly detection.


\vspace{-0.05in}
\section{Financial Application: User Study}
\label{sec:ustudy}

\subsection{User Study Setup}

\noindent 
{\bf Participants:~}
We recruited three professional fraud analysts from Capital One bank to participate in our user study. The analysts had years of experience respectively, with card fraud, bank fraud, and AML (anti-money laundering) as part of their job. We provided each analyst with an educational training of \method and its GUI on a demo dataset, explaining and demonstrating its particular functionalities and usage. User studies were conducted over Zoom with screen-sharing and recorded for later measurement (see metrics).

\noindent 
{\bf Data:~} 
We conducted the user studies on two separate datasets with ground-truth anomalies.
First dataset \fin is based on our simulator. From the
1999 Czech Financial Dataset\footnote{\url{https://www.kaggle.com/datasets/mariammariamr/1999-czech-financial-dataset}}, we obtained  
10,000 points via random sampling to which we fit a VAE.
Using the trained VAE and following Algo. \ref{alg:cap}, 
we simulated 1,000 normal points and 3 anomalous clusters with 20 anomalies each, based on different inflated feature subspaces.
Data contains  2 numerical and 4 categorical features, 
as well as the inlier or anomaly cluster labels. 

Second dataset \card 
contains a random sample of Capital One credit card transactions for a specific vendor over a period of time when they experienced a high (attempted) fraud rate. The credit card data has been anonymized with features renamed, numeric values renormalized, and categorical values hash-encoded. 
The dataset consists of 374 transactions, 82 of which are fraudulent or attempted fraudulent transactions. Each transaction has 3 numeric  and 4 categorical features, and a label indicating whether the transaction is normal or a fraudulent one. 

\noindent 
{\bf Procedure:~} We quantitatively evaluate the effectiveness and time-efficiency that \method provides via case studies. We also conduct an interview study and report qualitative feedback from the analysts.

\textbf{\em Case Studies:} 
We conducted three case studies on \fin and two case studies on \card. Each study associated with a specific task that an analyst is instructed to perform. 
To avoid leakage or prior familiarity between tasks, 
we used a separate one of three anomalous clusters in \fin for each task. \card came equipped with an existing domain-rule from earlier investigation, which helped prevent this issue.
We describe the different tasks as follows.

\vspace{-0.05in}
\begin{description}[font=$\bullet$\scshape\bfseries,leftmargin=*]
\item \textit{Task} 1: 
On \fin, we first ask the analyst to 
write a rule 1.a) using self-tools\footnote{Our analysts across various fraud domains used various ad-hoc tools such as Excel, pivot tables, SQL, etc. In contrast, \method proved to be a unified tool for all.}, and then 1.b) using \method. 
We measure and compare i) quality (coverage C and purity P) of rules
as well as ii) time it takes to write them (denoted time-to-rule).
On \card, we skip step 1.a) as the dataset readily came with an associated domain rule, developed  by earlier investigators.
\textit{This study is to quantify the added benefit of \method vs. ad-hoc tools that analysts otherwise use.   }

\item \textit{Task} 2: 
On \fin, we ask the analysts to explore 
   the automatically generated rules 
   and
    improve one candidate rule of their choice using the exploration (\expl) and the rule design interface (\rdi).
We measure how quickly and how much they can improve the rules, if at all, in terms of average C and P. 
On \card, the rule to be improved is specifically initialized as the readily available  domain-rule. 
\textit{This study is to quantify the added benefit of~ \expl and \rdi in improving a potentially suboptimal rule}.

\item \textit{Task} 3: 
  We ask the analyst to write rules solely using \expl and \rdi. 
We measure i) proximity of analyst-generated rules to the ground-truth, and ii) time it takes to write them (compared to avg. analyst time-to-rule via self-tools). 
We do \textbf{not} provide the analysts with any candidate rules (by simply setting C and P both to 1 in \cand s.t. no rules show up)---as those may readily include the ``ideal'' rule and obviate the study.
We conduct this study only on \fin, as for \card it would be the same as case 1.b) in Task 1 without permission to see candidate rules.
\textit{This study is to measure \method's role in helping the analyst get to the ideal rule.}
\end{description}
\vspace{-0.05in}

\noindent
\textbf{Metrics:~} We measure the  coverage (C) and purity (P) of the rules designed as well as the duration or time-to-rule in all case studies.  

\textbf{\em Interview Study:}
We followed the studies with a list of interview questions 
(see Sec. \ref{sssec:interviewqs}) to which the analysts responded with short answers. The interview probed for their feedback regarding the usability and functionality of \method. 

\subsection{Case Study Results}
\label{ssec:cases}

Tables \ref{tab:fin} and \ref{tab:card} in Appx. \ref{ssec:ustudyappx}, respectively for \fin and \card datasets, give the detailed results across tasks and analysts. This section summarizes the outcomes and take-aways for each task.

\begin{figure}[!ht]
\vspace{-0.125in}
\includegraphics[width=0.95\linewidth]{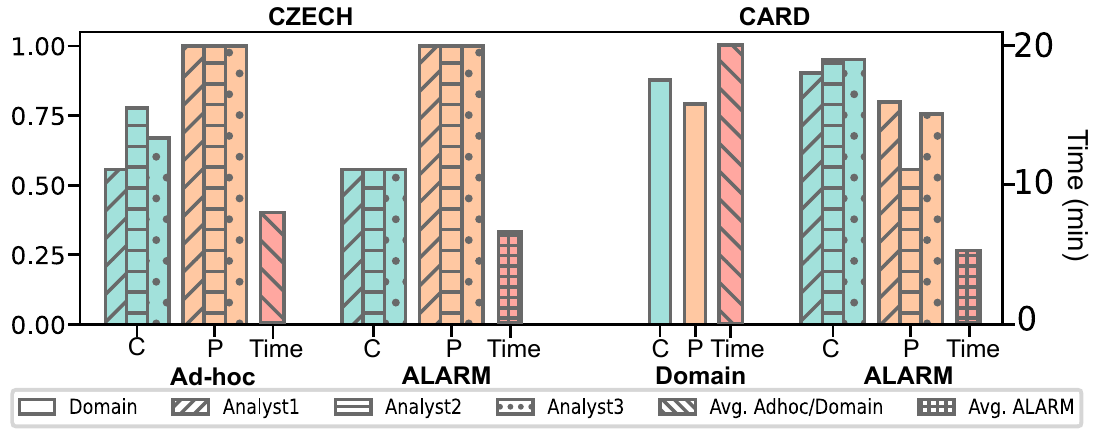}
   \vspace{-0.15in}
\caption{Task 1 contrasts rules by ad-hoc tools (\fin) or the domain-rule (\card) vs. \method-based rules across analysts.}
\vspace{-0.1in}
\label{fig:task1}
\end{figure}

 \textbf{Fig. \ref{fig:task1} for Task 1 (ad-hoc tools vs. \method) demonstrates that the analysts using \method generally produced comparable rules to those using ad-hoc tools or to the pre-existing domain rule.} On \fin, analysts have generated higher coverage rules using ad-hoc tools, only by having a disjunctive ``OR'' clause that treats the anomalies as two groups. On \card, all analysts consistently produced rules with higher coverage than that of the domain-rule, sacrificing purity slightly. Since the analysts target financial fraud, they typically prioritize high coverage to avoid potentially large monetary losses from false negatives.

\textbf{Furthermore, \method is more efficient.} Different from ad-hoc tools, \method's putting ``the components in one place, without having to switch between multiple pivot tables and Excel sheets'' provides an advantage. Average time-to-rule on \fin using \method (around 6 mins) is shorter than that the analysts spent using ad-hoc tools (8 mins). 
On \card, all analysts were quick (also 6 mins) to produce rules comparable to the domain-rule without any training, whereas the domain rules are created with significantly longer time (``about 10-30 mins'') and require specialized domain knowledge. As such, \method would be useful especially for the beginner analysts.

\begin{figure}[!ht]
\vspace{-0.125in}
\includegraphics[width=0.7\linewidth]{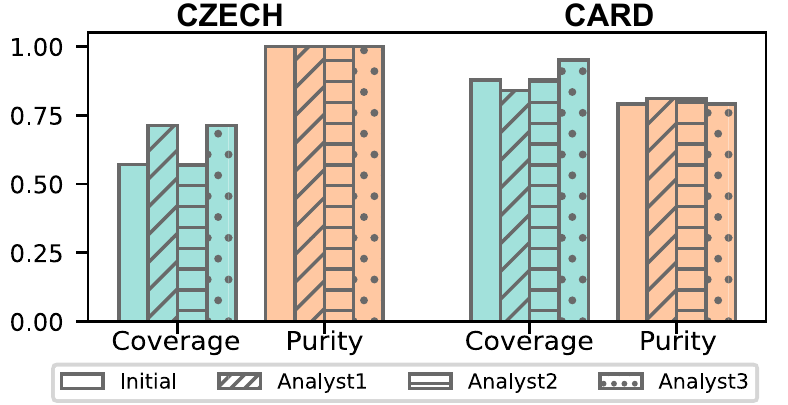}
   \vspace*{-0.1in}
\caption{Task 2 contrasts \method-based rules by the analysts vs. init. candidate-rule on \fin and domain-rule on \card.}
\vspace{-0.1in}
\label{fig:task2}
\end{figure}

\textbf{Fig. \ref{fig:task2} for Task 2 (initial-rule vs. improve-w/\method) shows that \method's interactive exploration and rule design can assist in improving existing rules.} On \fin, all analysts selected the same initial rule from among the candidates due to its simplicity (single predicate), and 
two of the three were able to improve to higher coverage without changing purity, within 2 mins on average. On \card, all three analysts have improved the purity of the domain-rule, with only one having to sacrifice coverage slightly, in about 6 mins on average. 
The improvements on \card are particularly notable, since the domain-rule for \card is already a carefully-crafted \textit{deployed} rule. 

\textbf{On Task 3 (ground-truth vs. \method-w/out-\cand) we find that the \method-based rules explored by the analysts and the (hidden) ground-truth rule are consistent with each other}, regarding their common usage of a predicate that aligns with the crucial ground-truth predicate ({\tt balance} between 60000 and 75000). 

\begin{wrapfigure}{ht!}{0.23\textwidth}
\vspace{-0.15in}
\begin{center}
\hspace{-0.8in}
    \includegraphics[width=0.275\textwidth]{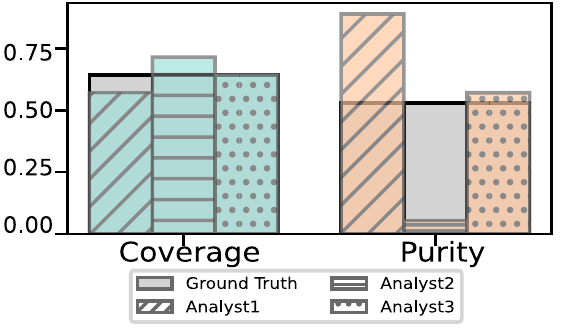}
    \vspace{-.125in}
\hspace{-0.7in}  \caption{Task 3: \method-based rules by analysts vs. ground-truth rule on \fin.\label{fig:task3}}
\end{center}
\vspace{-0.2in}
\end{wrapfigure}
As shown in Fig. \ref{fig:task3}, while one analyst built a very similar rule to the ground-truth w.r.t. coverage and purity, the others traded those in opposite directions\footnote{As automatic candidates were not allowed in Task 3, Analyst2, who had found them  most useful previously, started with a large trade-off and chose to focus on coverage.} by choosing different predicates besides {\tt balance}.  
The study showcased the space of alternative rules that \method allowed the analysts to explore, identify, and choose from based on potentially other, hard-to-quantify metrics such as policy, ethics, and deployment cost.




\vspace{-0.1in}
\subsection{Discussion on Lessons Learned}
\label{ssec:interview}
\vspace{-0.025in}

Based on analysts' feedback during the case studies and their answers to the interview questions, we compile a list of learned lessons as follows.
Overall, the analysts found \method to be 
``\textit{valuable for exploring the data and anomalies}'',
``\textit{useful for comparing and contrasting}'', and helpful in
 identifying ``\textit{specific pockets of risk}''. 

\noindent 
{\bf On efficiency:~} Analysts agreed that \method 
is ``\textit{a large time saver}'', and enjoyed that it 
allowed them to ``\textit{instantly generate rule candidates}'', 
``\textit{quickly adjust thresholds and calculate how these adjustments affected the coverage and purity}'', as well as
``\textit{quickly getting a sense of where the anomalies are and how they're spread/clustered}''.
 
\noindent
{\bf On automation \& interaction:}
While some analysts found \cand, i.e. 
``\textit{automatic rule mining component to be the most useful}'',
others perceived it as ``\textit{unable to generate optimal rules}'' which made them ``\textit{reluctant to trust}'' it ``\textit{in favor of writing [their] own}''. \rdi was unanimously valued both in terms of the efficacy and efficiency it provided over manual practice: 
``\textit{attempt to iterate ... was a massive value proposition as compared to doing this manually}.''

\noindent
{\bf On complexity:~} All analysts consistently took most advantage of the simple 
histogram and density plots, and some also the ``\textit{string}'' (i.e. parallel) plot to ``\textit{quickly and easily identify where anomalies were located}''.
However, 
MDS based summary viz. (esp. the axes) and the LookOut were deemed ``\textit{too complex to get into}''; suggesting that 
``\textit{individuals less familiar with ML may need robust setup instructions in order to understand and use [those]}''.

\noindent
\textbf{Other desired functionalities:~}
All analysts inquired if they could use 
multiple values or intervals per predicate, mimicking the  disjunctive ``OR''.
Several commented that the ``\textit{ability of the user to choose their own clusters}'' (also split or merge existing clusters) would be ``\textit{a powerful part of this tool}''.
We note that splitting a cluster would in fact mimic an ``OR''-based rule.
While designing their rule, they liked to observe ``\textit{highlights on visuals of the regions covered by rule}''.
One analyst suggested to add the flexibility to import additional data (e.g. last month's records from a specific vendor) while another suggested the ability to add new, analyst-crafted features.
Broadly, all analysts were eager to ``\textit{experiment with the tool using live data in [their] respective field}''  based on which they could 
 ``\textit{make more specific suggestions for  improvement}''.



\vspace{-0.05in}
\section{Related Work}
\label{sec:related}


\subsection{Anomaly Explanation}
While anomaly detection has a vast literature, anomaly explanation has attracted attention only recently. (See Panjei {\em et al.}'s survey \cite{panjei2022survey}.) 
Most existing techniques identify outlying features wherein the anomalies stand out the most, individually or in groups. 
Majority 
are model-agnostic and follow post hoc explanation strategies. 


In principle, 
strategies from supervised explainable ML 
can be used to explain the anomalies. 
One can first fit a regressor onto anomaly scores as the target variable, 
and use existing explanation methods for supervised models (see \cite{burkart2021survey,carrillo2021individual}). 
An issue with this approach relate to the intermediate step of model fitting and the fidelity of its fit to the anomaly scores.

A perhaps more critical 
issue with explanation methods 
is that they 
often disagree in terms of the explanations they output \cite{krishna2022disagreement}. While model-specific explanations may be preferable to model-agnostic ones due to fidelity, a study from the practitioners' perspective finds that they strongly prefer interaction 
and ``dialog'', rather than one-time, static explanations 
 \cite{lakkaraju2022rethinking}.
Our work constitutes a step forward toward the realm of interactive explainable systems.

\vspace{-0.1in}
\subsection{Visual Analytics for Anomaly Mining}
Over the decades, many visualization tools have been developed for  anomalies of various data types \cite{shi2020visual}, including spatial data \cite{cao2017voila,liao2010GPS}, graph and network anomalies \cite{goodall2018situ}, and multivariate-times series \cite{liu2021MTV}. These methods employ conventional visual analytic tools such as histograms and heatmaps, as well as innovative tools, however, they primarily aim to facilitate the \textit{discovery} of anomalies via human perception. As such, they often fall short in providing additional support beyond visualization, such as assisting analysts in taking action based on the anomalies detected. Many visualization tools are also domain-specific, focused on e.g. financial transaction monitoring \cite{leite2017eva,ko2016survey}, rumors and harmful bots on social media \cite{cao2015targetvue,cao2014fluxflow}, cyber-network surveillance \cite{mckenna2016cyber,goodall2009viassist,shiravi2011survey}, disease detection \cite{shneiderman2013improving},  traffic \cite{santhosh2020anomaly,riveiro2009reasoning}, and cloud-compute monitoring  \cite{xu2019clouddet}, etc. 

In contrast, \method is domain-agnostic and can be used with any mixed-typed point-cloud data. It focuses specifically on anomaly reasoning and rule design for future deployment. At the same time, \method aligns with human-in-the-loop anomaly discovery \cite{lamba2019learning,chai2020human,navarro2020hurra,smits2022panda} that harness human expertise to enhance the detection capabilities of ML algorithms.
\method is similar to studies that explain detected anomalies \cite{montambault2022pixal,ooge2022explain}, yet it further offers a comprehensive pipeline that covers the entire process from detection to human-in-the-loop reasoning and action-taking.









\section{Conclusion}
\label{sec:conclusion}




We presented \method, a new framework for end-to-end anomaly mining, reasoning and management that supports the human analyst in the loop. It offers unsupervised detection,  
anomaly explanations and an interactive GUI that guides analysts toward action, i.e. new rule design. User studies with fraud analysts validate \method's efficacy in finance, yet it can apply to combating 
 emerging threats in many other domains, thus is
open-sourced\textsuperscript{\ref{note1}} for the community.

\hide{
We presented \method, a new scalable open-source tool for distributed outlier detection (OD).
We described its design principles and the underlying distributed/data-parallel algorithms for shared-nothing cloud-computing platforms, and open-sourced its Apache PySpark implementation  at \la{\url{https://tinyurl.com/sparx2022}}.
    OD finds numerous applications, yet there are limited public-domain resources for \textit{distributed} OD as the vast majority of the literature focuses on single-machine algorithmic problems. 
Through extensive experiments, we showed that the few existing open-source tools do not match up with \method; they either do not scale well with dataset size or increasing dimensionality. Distinctly, \method is fully data-parallel, and scales linearly. 
We believe \method sets the state-of-the-art in terms of detection performance and scalability for distributed OD tasks. We expect it to increase the usability of OD on large-scale modern-day datasets that are already cloud-resident, and to offer significant impact in the applied domain for various business, engineering and scientific use cases.
}



\clearpage

\bibliographystyle{ACM-Reference-Format}
\bibliography{BIB/refs}

\clearpage

\appendix 
\section{Appendix}
\label{sec:appendix}

\hide{
\subsection{Anomaly Detection with \ex: Review}
\label{ssec:xstream_appx}

\ex \cite{manzoor2018xstream} is a  high-dimensional streaming data outlier detection algorithm, which consists of three main steps. First, it creates efficient data sketches to tackle high-dimensionality. 
Then, it builds efficient counting data structures for histogram-based density estimation in random feature subspaces. Finally, it scores
outlierness by the approximate density estimates.
We briefly describe each of these steps as follows.

\vspace{-0.05in}
\subsubsection{\bf Step 1. Data Projection}

Consider mixed-type data points $\bx_i \in \R^D$ where $D$ 
is the dimensionality upon one-hot-encoding (OHE) 
the categorical features. 
\ex creates a low-dimensional sketch/embedding $\bs_i$  for each point 
via random projections \cite{indyk1998approximate,achlioptas2003database}: 
\beq
\label{eqappx:sketch}
\bs_i = (\bx_i^T \br_1, \dots, \bx_i^T\br_K)
\eeq
where $\{\br_1,\ldots,\br_K\}$
depict $K$ random Gaussian vectors \cite{indyk1998approximate} or \textit{sparse} random vectors where with probability 1/3, $\br_k[F]\in \{\pm1\}$ and zero otherwise \cite{achlioptas2003database}.  
The latter choice not only is ``database-friendly'', i.e. more space and compute efficient, but can be also advantageous for outlier detection by effectively looking at data subspaces, reducing the masking effect of irrelevant features \cite{zimek2012survey}.

Notice that the same $\br_k \in \R^D$'s are used for all points over a stream, and hence need to be cashed.
However, for evolving streams wherein new features may emerge, $D$ is not fixed and in fact may be unknown apriori.
Then, the idea is not to cash, but to hash. Specifically, entries of each $\br_k$ is to be computed \textit{on-the-fly} via hashing, such that Eq. \eqref{eqappx:sketch} is rewritten as follows.
\beq
\label{eqappx:hash}
\scalemath{0.85}{
\bs_i[k] =  \sum_{F \in \mF_r} h_k(F) \;\cdot\; \bx_i[F]  
\;+ \sum_{F \in \mF_c}  h_k(F \oplus \bx_i[F]) \;\cdot\; 1
\;,\;\; k=1\ldots K
}
\eeq
\vspace{-0.1in}

\noindent
where $h_k(\cdot)$ is a hash function, $\mF_r$ and $\mF_c$ respectively denote the set of real-valued and categorical features, $\bx_i[F]$ is point $i$'s value of feature $F$, and $\oplus$ denotes the string-concatenation operator. 
Each hash function takes as input a string and returns $+1, -1$ or $0$ with respective probabilities $1/6, 1/6$ and 2/3.
(See \cite{manzoor2018xstream} for implementation details of such hash families.) 
For numerical features the input string is the feature name. For categorical features, it is the concatenation of the name and the corresponding feature value (both strings).
In effect, the sparse projection vector entries are computed  via hashing, i.e. $\br_k[F]=h_k[F]$, and multiplied with the corresponding feature value. For categorical features, the concatenated string corresponds to the OHE feature name with value $1$.

When triplet updates $<ID, F, \delta>$ arrive over the stream, where $\delta =\;$\texttt{\small{old\_val:new\_val}} for categorical features, the sketch can be updated by
\begin{equation}
\label{eqappx:update}
\hspace{-0.125in}
\scalemath{0.9}{
\bs_{ID}[k] = \begin{cases}
 \bs_{ID}[k] + h_k(F) \cdot \delta   {\text{\quad if real-valued $F$,}}
\\
\bs_{ID}[k] - h_k(F\oplus\text{\texttt{\small{old\_val}}}) + h_k(F\oplus\text{\texttt{\small{new\_val}}})   \text{  o.w.} 
\end{cases}
}
\end{equation}
for $k=1\ldots K$ such that $h_k(F\oplus\text{\texttt{\small{old\_val}}})$ returns zero when \texttt{\small{old\_val}} is \texttt{\small{null}}.
It is important to notice that Eq. \eqref{eqappx:update} can seamlessly handle a newly emerging feature $F$ that has never been seen before.

\subsubsection{\bf Step 2. Denstiy Estimation with Half-space Chains}

Anomaly detection relies on density estimation at multiple scales via a set of so-called Half-space Chains (HC), a data structure akin to multi-granular subspace histograms.
Each HC has a length $L$ (or $L$ layers), along which the (projected) feature space $\mF_p$ is recursively halved on a randomly sampled (with replacement) feature, where $f_l \in \{1,\ldots,K\}$ denotes the feature at level $l=1,\ldots, L$.
As such, a point can reside in one of 2 bins at level 1, one of 4 bins at level 2, and in general one of $2^l$ bins at level $l$.
Given the sketch $\bs$ of a point, the goal is to efficiently identify the bin it falls into at each level.

Let $\bdel \in \R^K$ be the vector of initial bin widths, which is equal to half
the range of the projected data along each dimension $f \in \mF_p$.
Let $\bzb_l \in \Z^K$ denote the bin identifier of $\bs$ at level $l$, initialized to all zeros.
At level 1, the bin-id is updated as $\bzb_1[f_1] = \lfloor \bs[f_1]/\bdel[f_1] \rfloor$.
In general, the bin-id at consecutive levels can be computed \textit{incrementally}, following

\vspace{-0.15in}
\begin{equation}
\label{eqappx:binid}
\scalemath{0.9}{
\bzb_l[f_l] = \lfloor \bz_l[f_l] \rfloor \;\text{  s.t.  } \;
\bz_l[f_l] = \begin{cases}
\bs[f_l]/\bdel[f_l]  \text{\quad if $o(f_l,l) =1$, and}
\\
2\bz_l[f_l] \text{\quad \quad\;\; o.w.; if $o(f_l,l)> 1$ } 
\end{cases}
}
\end{equation}
where $o(f_l,l)$ denotes the number
of times feature $f_l = \{1,\ldots,K\}$ has been sampled in the chain until and
including level $l$.
We note that a small uniformly random value $\epsilon_l \in (0, \bdel[f_l])$, called shift, is added to the sketches at each level to remedy issues for nearby points around fixed bin boundaries. We omit those for brevity and refer to \cite{manzoor2018xstream} for details.

Notice that all points with the same unique $\bzb_l$ reside in the same histogram bin at level $l$. As such, level-wise (multi-scale) densities are to be estimated by counting the number of points with the same bin-id at each level. This can be done by a dictionary (or perfect hash) data structure.  
The number of possible bins, however, grows exponentally with $l$.
Even though data is not necessarily spread to all bins, number of non-empty bins (and hence the size of the dictionary) can grow very large for large $L$. Instead, \ex obtains approximate counts via a count-min-sketch \cite{cormode2005improved}, the size of which is user-specified, i.e. constant.

Overall, \ex is an ensemble of $M$ Half-Space Chains, $\mH = \{HC^{(m)} := (\bdel, \bfl^{(m)}, \be^{(m)},  \mC^{(m)})\}_{m=1}^M$ where each HC is associated with the following list of meta-data;
 (i) the bin-width per feature $\bdel \in \R^K$,
 (ii) the sampled feature per level $\bfl^{(m)} \in \Z^L$,
(iii) the random shift value per level $\be^{(m)} \in \R^L$, and (iv) the counting data structure per level $\mC^{(m)} = \{C_l^{(m)}\}_{l=1}^L$.

\subsubsection{\bf Step 3. Outlier Scoring}
To score a given (updated) sketch for outlierness, the count of points in the bin that it falls into is identified at each level $l$ of a HC, denoted $C_l^{(HC)}[\bzb_l]$.
The count is extrapolated via multiplying by $2^l$ to estimate the total count if the data were distributed uniformly. Smallest estimate across levels\footnote{Note that the counts can be compared across levels after extrapolation.} is taken as the outlier score, as 

\vspace{-0.15in}
\beq
\label{eqappx:score}
\scalemath{0.9}{
O^{(m)}(\bs) = 
\min_l \; 2^l \cdot C_l^{(m)}[\bzb_l] \;.
}
\eeq
The scores are averaged across all HCs to obtain the final outlier score.
A lower value indicates 
a granularity at which the point resides in a relatively sparse region, and hence 
higher outlierness. 
}

\begin{table*}[!th]
\caption{AUROC performances produced by \ex and Isolation Forest (IF). Both IF+SHAP and DIFFI utilize Isolation Forest, while \ex + SHAP and \ex (with and without projectinos) utilize \ex.}
\vspace{-0.13in}
\begin{tabular}{r c c c c c c}
\hline
{Method} & {Seismic} & {KDDCUP} & {Hypothyroid} & {Cardio} & {Satellite} & {BreastW} \\
\hline
Isolation Forest & 
0.998$\pm$0.001 &
0.992$\pm$0.000 & 
0.982$\pm$0.002 & 
0.999$\pm$0.001 & 
0.997$\pm$0.000 &  
1.000$\pm$0.001 \\
\ex No Proj. &
0.993$\pm$0.002 & 
0.963$\pm$0.005 & 
0.995$\pm$0.002 & 
0.998$\pm$0.002 & 
1.000$\pm$0.003 & 
0.999$\pm$0.001 \\

\ex Proj. =15&
0.968$\pm$0.008  &
0.932$\pm$0.007&
0.732$\pm$0.010&
0.998$\pm$0.002&
0.991$\pm$0.002&
0.998$\pm$0.003\\

\ex Proj. = 20& 
0.972$\pm$0.002  &
0.945$\pm$0.005& 
0.873$\pm$0.007& 
0.998$\pm$0.002 &
0.992$\pm$0.006 &
0.998$\pm$0.002 \\

\ex Proj. =30 &
0.971$\pm$0.005 &
0.951$\pm$0.004&
0.919$\pm$0.005&
0.998$\pm$0.003&
0.998$\pm$0.003&
0.999$\pm$0.002 \\
\hline
\end{tabular}
\label{tab:auroc}
\end{table*}

\subsection{Details on Explanation Evaluation}
\label{ssec:expeval}

\subsubsection{Inflating Anomalies}
\label{sssec:inflate}
In Steinbruss \textit{et al.} \cite{steinbuss2021benchmarking}, anomalies are categorized into several types, including local and global anomalies. Local anomalies refer to anomalies that occur in the vicinity of local neighborhoods, while global anomalies are dispersed throughout the entire feature space and are vastly different from normal data.

For real-valued features, the marginal distribution of the data is modeled first using Gaussian Mixture Models (GMM) \cite{milligan1985}. In our experiment, we select the optimal number of clusters $G$ with the best BIC measure. The  corresponding means $\mu_i \in \mathbb{R}$, variances $\sigma_i \in \mathbb{R}$, and mixing proportions  $\pi_i \in [0,1]$ are found, for each $i \in G$. To generate local anomalies, the variance is multiplied by $\alpha = 3$ and anomalies are identified using the same GMM with the modified variances $\alpha*\sigma_i$. These anomalies are close to normal points but fall outside the cluster. For global anomalies, the minimum and maximum of the data points are found, and instances are generated from a uniform distribution \cite{maitra2010simulating} with an extended minimum and maximum, multiplied by $\beta = 1.2$. These anomalies come from a different distribution and are scattered throughout the entire feature space. For categorical-valued features, we do not distinguish between local and global anomalies. Instead, we estimate the empirical categorical distribution of the feature's values and identify which class values are unlikely based on the observed distribution. Anomalies are created by switching from a common class value to a low-probability one. In our study, we selected anomaly features with the fewest occurrences.

\subsubsection{Setup Details}

\textbf{Hyperparameter configurations.~}

For generation of synthesis data, we use all the normal data, removing the missing features and features with only one value. We convert all categorical features into one-hot encoding and normalize all continuous features using min-max normalization. The total of dataset dimensions are described in Table \ref{tab: datasetdim}. During generation, the generated categorical feature values are selected with the maximum-probability and then decoded. Our synthesizing VAE currently only supports a Gaussian prior, but it can also be extended into multi-stage VAEs or VAEs with various priors \cite{jang2016gumbel, storkey2018vaevamp}, which better capture the categorical distributions. 

\begin{wraptable}{r}{3cm}
\vspace{-0.15in}
\centering
\caption{Training Dataset dimensions. }
\vspace{-0.15in}
	\centering
	\hspace{-0.25in}
	{\scalebox{0.8}{
    \begin{tabular}{lr}
    \toprule
    \textbf{Name} &  \textbf{Total Dim} \\
    \midrule
    Seismic &  $2414 \times 21$ \\
    KDDCUP &  $15000 \times 37$ \\
    Hypothyroid &  $1878 \times 30$ \\
    Cardio &  $1831 \times 21$ \\
    Satellite &  $4399 \times 36$ \\
    BreastW & $683 \times 9$ \\
    \bottomrule
    \end{tabular}
}}
\vspace{-0.1in}
    \label{tab: datasetdim}
\end{wraptable} 

The VAE's hyperparameters are given as the following: learning rate is $5e^{-4}$, iterations is 1500 epochs, and batch size is 256. VAE has three encoder layers with width $[500, 200, 30]$, respectively, and ReLU activation units. The hidden dimension is 15, so the last layer outputs the mean and variance. After sampling, we feed the hidden vector into decoder with three layers and ReLU activation units, and width are $[200,500, \text{output size} *2]$. The output of the VAE is the posterior mean and variance, while reconstructed samples are generated with the updated mean and variance. For generation of the anomalies, we set threshold variable $\epsilon = 0.5$, and force generated anomalies to show in a lower-probability region.

For Isolation Forest (IF), we use number of estimators equals 100, max samples equals 256 and set contamination rate at 0.1. To enable detection on mixed-type data, all categorical features are converted into one-hot-encoding and all real-valued features are min-max normalized to between 0 and 1. Isolation Forest's hyperparmaters are utilzied by both IF+SHAP and DIFFI, to keep the results comparable. 

For \ex, we fix number of Half-space Chains $M$$=$$200$, with each chain length 
 $L$$=$$20$. In our experiment for feature importances, we do not use the CMS hashtable as the structure of \ex Half-space Chains, since only static datapoints are tested at this time. Each Half-space Chain is constructed with python's list structure. However, if using the distributed version of \ex (Sparx \cite{sparx22}) with CMS hashtable, CMS hashtable size is usually set at $m = 10$, with number of rows $r = 3$ and number of columns $c = 50$. The same \ex parameters are used for \ex+SHAP and pure \ex, so the explanation method's results are comparable.

 For DIFFI, no additional parameters are required. For SHAP, a sped-up TreeExplainer is provided based on  \cite{lundberg2020local}. We repeat each experiment three times on CPU configuration: AMD EPYC 7502 with 32 cores, and record the average and standard deviation (std) of AUROC, time, and NDCG scores. 

\subsubsection{Detection Performances}
\label{sssec:perf}
Table \ref{tab:auroc} gives the performances results of the detection methods.

\subsection{Details on User Study}
\label{ssec:ustudyappx}
For each task, we track the time spent by the analysts, starting from the beginning of their investigation until they determine that they have discovered the rules. After the rules are found, we document the coverage and purity of the rules on grouped anomalies and also keep a record of the specifics of the discovered rules. Before the start of three tasks, the analysts are given an example interface of \method, where each component of interactive visual interface is shown using a synthetic dataset and its functionality is explained to the analysts. However, the analysts can also click the "?" button on interactive visual interface to read about the component's information. The analysts can also ask questions of \method. The question time will not be counted towards the final time.

For \fin dataset, we divide the detected anomalies into three clusters, so each of the task will be given a different cluster (shown in \clus). \card dataset is a single-cluster of anomalous transactions, and it has already came with a set of existing domain rules, developed by earlier investigators. To prevent duplication of work by analysts, we limit the use of \card to only Task 1 (for creating rules with \method) and Task 2 (for enhancing \method's existing rules). 

In Task 1, analysts are presented with a cluster of anomalies that have either $0$ (normal) or $1$ (anomaly) classification labels. The labels are either specified by domain rules (\card dataset) or predicted by the \ex (\fin dataset). The analysts are then allowed to investigate these anomalies using their own preferred tools. All three analysts chose to use Excel. With Excel, they ranked columns and used pivot tables to identify anomaly patterns. For \card, which already has an established domain rule, the analysts only need to utilize the \method to uncover the anomaly patterns. The analysts are allowed to type-in any coverage and purity thresholds, and utilize the existing rule candidates. They are also allowed to utilize exploration panel (\expl) and rule design interface (\rdi) to improve existing rules or come up with new rules. For \fin, Analyst 1 and Analyst 2 both wrote down a set of rules that are comparable in terms of coverage and purity in 4 and 3 minutes, respectively. However, they commented that they wanted to find additional rules because the ones they found are too simple, with only one predicate. They spent an additional 5 and 13 minutes examining and developing a second set of rules. Only the time with the first rules, coverage, and purity generated by the first set of rules are taken into account for fairness, but all the outcomes are documented in the table. For \card, Analyst 3 also discovered two sets of rules. The time, coverage, and purity of the first set of rules are used, while the second set are also documented in the table.

In Task 2 for \fin, we use a different anomaly cluster and the analysts are presented with three potential sets of rules (with a coverage threshold set to 0.5 and a purity threshold set to 0.5) generated by x-PACS in the rule candidate view panel (\cand). The analysts all began by examining the middle set. The middle set of rules has a balanced combination of high coverage and purity, and only contains one simple predicate that can be easily modified. All three analysts started with experimenting with different feature combinations on the exploration panel (\expl). Then, they added or modified existing rules in the rule design interface (\rdi), and saw the resulting changes of coverage and purity. The final rules were recorded when the analysts reached a conclusion that the selected rules are their final choices. For \card, the analysts were given one set of pre-existing domain rules of the same anomaly group as in Task 1, while the rest of the process remains similar.

Task 3 is only carried out with the \fin dataset using a different cluster of anomalies compared to the previous two tasks. The analysts do not have access to the "ideal" rules that best explain the group of anomalies. They are allowed to use the rule design interface (\rdi) and exploration panel (\expl) to investigate the properties of the anomalies. All three analysts explored various features for plots on the exploration panel (\expl). Analyst 1 first came up with two rules with lower coverage ($0.071$, $0.286$) and then decided to continue looking for rules with higher coverage. All three analysts quickly developed simple rules with high coverage ($0.643$, $0.714$, and $0.643$, respectively) but low purity ($0.086$, $0.055$, and $0.106$, respectively). Analyst 2 noted that such low purity ($0.055$) is acceptable if a limited number of inliers need to be manually examined, while the other two analysts were not content with the purity and continued their search. Analyst 1 found another set of rules with high coverage ($0.571$) and purity ($0.889$) in just 5 minutes. Meanwhile, Analyst 3 aimed to improve the coverage without significantly sacrificing the purity. After two attempts, Analyst 3 successfully created rules with good coverage ($0.643$) and purity ($0.562$). It took a total of around 8 minutes for Analyst 3 to come up with three sets of rules. The final sets of rules, as they believed that further improvements were not possible, were considered in our analysis.

\subsubsection{Detailed Results}
Table \ref{tab:fin} and Table \ref{tab:card} provide the user study results on \fin and \card datasets.

\begin{table*}[ht!]
\caption{User study results on individual (3) tasks with three fraud analysts on the \fin dataset. Reported are Coverage (C), Purity (P), Time required (when applicable) and associated Rule(s).}
\vspace{-0.05in}
\scalebox{0.95}{
\begin{tabular}{l|lllll|llll}
\toprule
\multirow{7}{*}{\textbf{Task 1}} & \multicolumn{5}{c|}{\textbf{Ad-hoc Tools}} & \multicolumn{4}{c}{\textbf{\method}} \\ \cline{2-10} 
 &  & \multicolumn{1}{c}{\textbf{C}} & \multicolumn{1}{c}{\textbf{P}} & \multicolumn{1}{c}{\textbf{Time}} & \multicolumn{1}{c|}{\textbf{Rules} } & \multicolumn{1}{c}{\textbf{C}} & \multicolumn{1}{c}{\textbf{P}} & \multicolumn{1}{c}{\textbf{Time}} & \multicolumn{1}{c}{\textbf{Rules} } \\ \cline{2-10} 
 & Analyst 1 & 0.556 & 1.000 & $8^\prime 07^{\prime\prime}$ & \begin{tabular}[t]{@{}l@{}} \tt{amount} $\in [5000,7000]$, \\ \tt{balance} $\in [90000,\infty)$ \end{tabular} & 0.556 & 1.000 & $3^\prime 08^{\prime\prime}$ & \tt{k\_symbol} = UVER \\
 &  &  &  &  & & 0.667 & 0.667 & $4^\prime 48^{\prime\prime}$ & \begin{tabular}[t]{@{}l@{}}\tt{amount} $\in [5501,6501]$, \\ \tt{balance} $\in[65475, 99807]$\end{tabular} \\\\
 
 & Analyst 2 & 0.777 & 1.000 & $7^\prime 09^{\prime\prime}$ & \begin{tabular}[t]{@{}l@{}} \tt{amount} $\in [39000,43000]$ \\ OR \tt{balance} $\in [90,000,\infty)$ \end{tabular} & 0.556 & 1.000 & $2^\prime 37^{\prime\prime}$ & \tt{balance} $\in [90475,\infty)$  \\
 & & &  &  &  & 0.556 & 1.000 & $13^\prime 28^{\prime\prime}$ & \begin{tabular}[t]{@{}l@{}}\tt{operation} = TFA, \\ \tt{amount} $\in [5390,6502]$, \\ \tt{k\_symbol} = UVER\end{tabular} \\\\
 
 & Analyst 3 & 0.556 & 1.000 & $5^\prime 30^{\prime\prime}$ & \begin{tabular}[t]{@{}l@{}}
 \tt{k\_symbol} = UVER \end{tabular} & 0.556 & 1.000 & $14^\prime 00^{\prime\prime}$ & \begin{tabular}[t]{@{}l@{}} \tt{type} = CREDIT, \\ \tt{operation} = TFA,\\ \tt{amount} $\in [5390,6502]$,\\ \tt{k\_symbol} = UVER\end{tabular} \\
 & & 0.666 & 1.000 & $8^\prime 32^{\prime\prime}$ & \begin{tabular}[t]{@{}l@{}}
 \tt{k\_symbol} = UVER\\
 OR \tt{k\_symbol} = POJISTNE,\\  \tt{operation} = TFA \end{tabular} &  &  &  &  \\
 
 \hline
\multirow{5}{*}{\textbf{Task 2}} & \multicolumn{5}{c|}{\textbf{Initial} pick from \cand} & \multicolumn{4}{c}{\textbf{Improve} w/ \method} \\ \cline{2-10} 
 &  & \multicolumn{1}{c}{\textbf{C}} & \multicolumn{1}{c}{\textbf{P}} & \multicolumn{2}{c|}{\textbf{Rules} } & \multicolumn{1}{c}{\textbf{C}} & \multicolumn{1}{c}{\textbf{P}} & \multicolumn{1}{c}{\textbf{Time}} & \multicolumn{1}{c}{\textbf{Rules} } \\ \cline{2-10} 
 & Analyst 1 & 0.571 & 1.000 & \multicolumn{2}{l|}{\tt{amount} $\in[ 42034,42849]$ } & 0.714 & 1.000 & $2^\prime 50 ^{\prime\prime}$ & \tt{amount} $\in [36001, 42849]$ \\\\
& Analyst 2 & 0.571 & 1.000 & \multicolumn{2}{l|}{\tt{amount} $\in[ 42034,42849]$ }& 0.571 & 1.000 & $7^\prime 50^{\prime\prime}$ & \tt{amount} $\in[42034, 42849]$ \\\\
 
 & Analyst 3 & 0.571 & 1.000 & \multicolumn{2}{l|}{\tt{amount} $\in[ 42034,42849]$ }& 0.714 & 1.000 & $1^\prime 48^{\prime\prime}$ & 
 \begin{tabular}[t]{@{}l@{}}\tt{amount} $\in [37001, 40475]$, \\ \tt{balance} $\in[ 40475, 60475]$\end{tabular}
  \\ \hline
\multirow{7}{*}{\textbf{Task 3}} & \multicolumn{5}{c|}{\textbf{Ground-truth rule}} & \multicolumn{4}{c}{\textbf{\method w/out \cand}} \\ \cline{2-10} 
 & \multicolumn{1}{c}{} & \multicolumn{1}{c}{\textbf{C}} & \multicolumn{1}{c}{\textbf{P}} & \multicolumn{2}{c|}{\textbf{Rules}} & \multicolumn{1}{c}{\textbf{C}} & \multicolumn{1}{c}{\textbf{P}} & \multicolumn{1}{c}{\textbf{Time}} & \multicolumn{1}{c}{\textbf{Rules}} \\ \cline{2-10} 
 & Analyst 1 & 0.643 & 0.529 & \multicolumn{2}{l|}{\begin{tabular}[t]{@{}l@{}}\tt{balance} $\in [60000,75000]$, \\ \tt{k\_symbol} = POJISTNE\end{tabular}} & 0.071 & 0.500 & $5^\prime 02 ^{\prime \prime}$ & \begin{tabular}[t]{@{}l@{}} \tt{type} = CREDIT, \\ \tt{operation} = TTA \end{tabular} \\

 &  &  & & \multicolumn{2}{l|}{} & 0.286 & 0.008 & $1^\prime 18 ^{\prime \prime}$ & \begin{tabular}[t]{@{}l@{}} \tt{type} = ISSUE, \\ \tt{operation} = TTA\end{tabular} \\

 &  &  & & \multicolumn{2}{l|}{} & 0.643 & 0.086 & $0^\prime 14 ^{\prime \prime}$ & \begin{tabular}[t]{@{}l@{}} \tt{operation} = TFA \end{tabular} \\
 
 & &  &  &  & & 0.571 & 0.889 & $4^\prime 59^{\prime\prime}$ & \begin{tabular}[t]{@{}l@{}} \tt{type} = ISSUE, \\ \tt{operation} = TFA, \\ \tt{balance} $\in [59475, \infty)$\end{tabular} \\ \\
 & Analyst 2 & same & same & \multicolumn{2}{l|}{\begin{tabular}[t]{@{}l@{}}same \\ \end{tabular}} & 0.714 & 0.055 & $3^\prime 17^{\prime\prime}$ & \tt{balance} $\in [50000,\infty)$ \\\\

 &  Analyst 3 & same & same & \multicolumn{2}{l|}{\begin{tabular}[t]{@{}l@{}}same \\ \end{tabular}} & 0.643 & 0.106 & $3^\prime 00^{\prime\prime}$ & \begin{tabular}[t]{@{}l@{}}\tt{balance} $\in [60475,80475]$, \\ \tt{amount} $\in [5001,42888]$ \end{tabular} \\ 
 &   &  &  &  & & 0.571 & 1.000 & $3^\prime17 ^{\prime\prime}$ & \begin{tabular}[t]{@{}l@{}} \tt{balance} $\in [59475,80475]$, \\ \tt{amount} $\in [5001,10001]$, \\ \tt{operation} = TFA \end{tabular}  \\
 &  &  &  & &  & 0.643 & 0.562 & $1^\prime13^{\prime\prime}$ & \begin{tabular}[t]{@{}l@{}}\tt{balance} $\in [59475,80475]$, \\ \tt{amount} $\in [5001,10001]$, \\ \tt{k\_symbol} = POJISTNE \end{tabular} \\ \bottomrule
\end{tabular}}
\label{tab:fin}
\end{table*}

\begin{table*}[ht!]
\caption{User study results on individual (2) tasks with three fraud analysts on the \card dataset. Reported are Coverage (C), Purity (P), Time required (when applicable) and associated Rule(s).}
\vspace{-0.05in}
\scalebox{1.0}{
\begin{tabular}{l|lllll|llll}
\toprule
\multirow{5}{*}{\textbf{Task 1}} & \multicolumn{5}{c|}{\textbf{Domain-rule}} & \multicolumn{4}{c}{\method} \\ \cline{2-10} 
 &  & \multicolumn{1}{c}{\textbf{C}} & \multicolumn{1}{c}{\textbf{P}} & \multicolumn{2}{c|}{\textbf{Rules} } & \multicolumn{1}{c}{\textbf{C}} & \multicolumn{1}{c}{\textbf{P}} & \multicolumn{1}{c}{\textbf{Time}} & \multicolumn{1}{c}{\textbf{Rules} } \\ \cline{2-10} 
 & Analyst 1 & 0.878 & 0.791 & \multicolumn{2}{l|}{\begin{tabular}[t]{@{}l@{}}\tt{feature2} $\in [110, \infty)$, \\ \tt{feature5} = SSDD, \\ \tt{feature6} $\in [350, \infty)$ \end{tabular}} & 0.902 & 0.799 &$3^\prime24^{\prime\prime}$ & \begin{tabular}[t]{@{}l@{}} \tt{feature2} $\in [70,\infty]$,  \\
 \tt{feature5} = SSDD, \\
 \tt{feature6} $\in [500, \infty)$ \end{tabular} \\\\
 
 & Analyst 2 & same & same & \multicolumn{2}{l|}{\begin{tabular}[c]{@{}l@{}} same \end{tabular}} & 0.951 & 0.557 & $6^\prime15^{\prime\prime}$ & \begin{tabular}[t]{@{}l@{}}\tt{feature3} = BBBBB, \\ \tt{feature4} = 1, \\ \tt{feature5}= SSDD\end{tabular} \\\\
 
 & Analyst 3 & same & same & \multicolumn{2}{l|}{\begin{tabular}[t]{@{}l@{}} same \end{tabular}} & 0.951 & 0.756 & $7^\prime 01^{\prime\prime} $& \begin{tabular}[t]{@{}l@{}}\tt{feature1} $\in[70,700]$, \\\tt{feature2} $\in [20,\infty)$, \\ \tt{feature4} = 1, \\ \tt{feature5} = SSDD, \\ \tt{feature6}
 $\in(320, \infty)$ \end{tabular}  \\ 
 & & & & & & 1.00 & 0.547 & $0^\prime24^{\prime\prime}$ & \begin{tabular}[t]{@{}l@{}}\tt{feature1} $\in [70,700]$, \\\tt{feature2} $\in [20,\infty)$, \\ \tt{feature4} = 1,  \\ \tt{feature6}  $\in [320, \infty)$ \end{tabular}\\
 \hline
\multirow{5}{*}{\textbf{Task 2}} & \multicolumn{5}{c|}{\textbf{Initial} set to \textbf{Domain-rule} } & \multicolumn{4}{c}{\textbf{Improve} w/ \method} \\ \cline{2-10} 
 &  & \multicolumn{1}{c}{\textbf{C}} & \multicolumn{1}{c}{\textbf{P}} & \multicolumn{2}{c|}{\textbf{Rules} } & \multicolumn{1}{c}{\textbf{C}} & \multicolumn{1}{c}{\textbf{P}} & \multicolumn{1}{c}{\textbf{Time}} & \multicolumn{1}{c}{\textbf{Rules} } \\ \cline{2-10} 
 & Analyst 1 & 0.878 & 0.791 & \multicolumn{2}{l|}{\begin{tabular}[t]{@{}l@{}}\tt{feature2} $\in [110, \infty)$, \\  \tt{feature5} = SSDD,\\  \tt{feature6} $\in [350, \infty)$ \end{tabular}} & \multicolumn{1}{r}{0.841} & 0.812 & $7^\prime 50^{\prime\prime}$ & \begin{tabular}[t]{@{}l@{}}\tt{feature2} $\in [110, \infty)$, \\ \tt{feature5} = SSDD, \\ \tt{feature6} $\in [500,\infty)$ \end{tabular} \\\\
 
 & Analyst 2 & same & same & \multicolumn{2}{l|}{\begin{tabular}[t]{@{}l@{}} same \end{tabular}} & 0.878 & 0.809 & $6^\prime 00 ^{\prime\prime}$& \begin{tabular}[t]{@{}l@{}} \tt{feature2} $[110, \infty)$, \\ \tt{feature4} = 1, \\ \tt{feature5} = SSDD, \\ \tt{feature6} $[350, \infty)$ \end{tabular} \\\\
 
 & Analyst 3 & same & same & \multicolumn{2}{l|}{\begin{tabular}[t]{@{}l@{}}same \end{tabular}} & 0.915 & 0.798 & $4^\prime 36^{\prime\prime}$ & \begin{tabular}[t]{@{}l@{}} \tt{feature1} $\in [70, 680]$ \\ \tt{feature2} $\in [80, \infty)$, \\ \tt{feature4} = 1, \\ \tt{feature5} = SSDD, \\
 \tt{feature6} $[350, \infty)$
 \end{tabular}  \\ \bottomrule
\end{tabular}}
\label{tab:card}
\end{table*}

\subsubsection{Interview Study: Questions}
\label{sssec:interviewqs}

\ben
\item   What did you think of the \textit{anomaly summarization/clustering} component (\clus)? (How) was it useful?

\item  What did you think of the \textit{candidate rules} presented (\cand)? (How) were they useful?
\item  (How) were the \textit{interactive tools} provided on the interface useful?
\ben
      \item  Were you able to sufficiently \sloppy{\textit{inspect/explore} the \underline{detected anomalies}} (\expl)?
      \item  Were you able to sufficiently \textit{explore/refine} the \underline{rule(s)} (\rdi)?
      \een
\item In what ways did you find it most valuable/useful to \textit{improve current practice}?
\item What are your \textit{suggestions for improvement}?
\een



\end{document}